\documentclass[10pt,twocolumn,letterpaper]{article}

\usepackage[pagenumbers]{wacv} 

\usepackage{graphicx}
\usepackage{amsmath}
\usepackage{amssymb}
\usepackage{booktabs}

\usepackage[accsupp]{axessibility}
\usepackage[pagebackref,breaklinks,colorlinks]{hyperref}

\usepackage[capitalize]{cleveref}
\crefname{section}{Sec.}{Secs.}
\Crefname{section}{Section}{Sections}
\Crefname{table}{Table}{Tables}
\crefname{table}{Tab.}{Tabs.}

\newcommand{\method}{AnomalyDINO}
\def\hcolor{gray!20}

\usepackage{graphicx}
\usepackage{cleveref}
\usepackage{subcaption}     
\usepackage{multirow}
\usepackage[table,xcdraw]{xcolor}

\usepackage[export]{adjustbox}

\newcommand{\std}[1]{\tiny{$\pm$#1}}

\usepackage{tcolorbox}

\usepackage{calc}
\newcommand{\printtowidth}[2]{%
  \makebox[\widthof{#1}]{#2}%
}

\usepackage{enumitem,amssymb}
\usepackage{pifont}
\newcommand{\cmark}{\ding{51}}%
\newcommand{\xmark}{\ding{55}}%
\usepackage[normalem]{ulem}
\useunder{\uline}{\ul}{}
\newcommand{\grayuline}[1]{{\color{gray}\uline{{\color{black}#1}}}}

\begin{document}

\title{\method{}: 
Boosting Patch-based Few-shot Anomaly Detection \\ with DINOv2}

\author{\textbf{Simon Damm}$^1$, \textbf{Mike Laszkiewicz}$^1$, \textbf{Johannes Lederer}$^2$, \textbf{Asja Fischer}$^1$\\
$^1$Department of Computer Science, Ruhr University Bochum, Germany \\
$^2$Department of Mathematics, Computer Science, and Natural Sciences, \\ University of Hamburg, Germany \\
{\tt\small \{simon.damm, mike.laszkiewicz, asja.fischer\}@rub.de} \\
{\tt\small johannes.lederer@uni-hamburg.de}
}
\maketitle

\begin{abstract}
    Recent advances in multimodal foundation models have set new standards in few-shot anomaly detection.
    This paper explores whether high-quality visual features alone are sufficient to rival existing state-of-the-art vision-language models.
    We affirm this by adapting DINOv2 for one-shot and few-shot anomaly detection, with a focus on industrial applications.
    We show that this approach does not only rival existing techniques but can even outmatch them in many settings.
    Our proposed vision-only approach, \method, 
    follows the well-established patch-level deep nearest neighbor paradigm, and enables both image-level anomaly prediction and pixel-level anomaly segmentation.
    The approach is methodologically simple and training-free and, thus, does not require any additional data for fine-tuning or meta-learning.
    Despite its simplicity, \method{} achieves state-of-the-art results in one- and few-shot anomaly detection (e.g., pushing the one-shot performance on MVTec-AD from an AUROC of $93.1\%$ to $96.6\%$).
    The reduced overhead, coupled with its outstanding few-shot performance, makes \method{} a strong candidate for fast deployment, e.g., in industrial contexts.
\end{abstract}


\section{Introduction}
\label{sec:Intro}

\begingroup
    \renewcommand\thefootnote{}
    \footnotetext{
    This paper was accepted at the Winter Conference on Applications of Computer Vision (WACV), 2025. 
    The final version is available
    \href{https://openaccess.thecvf.com/content/WACV2025/html/Damm_AnomalyDINO_Boosting_Patch-Based_Few-Shot_Anomaly_Detection_with_DINOv2_WACV_2025_paper.html}{here} (Open Access) and on IEEE Xplore.}
\endgroup

\begin{figure*}[t]
    \centering
    \includegraphics[width=0.94\linewidth]{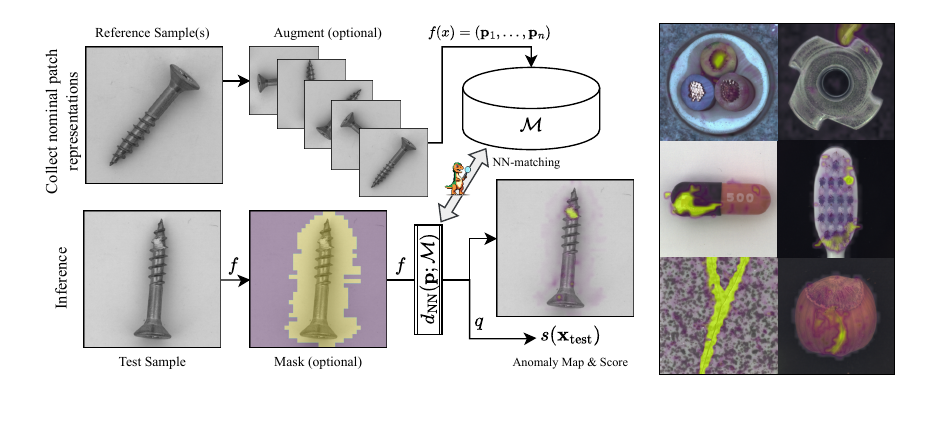}
    \vspace{-1cm}
    \caption{\textbf{Anomaly detection with AnomalyDINO} based on a single immaculate reference sample (here category `Screw' from MVTec-AD).
    We collect the nominal patch representations from the (potentially augmented) reference sample(s) in the memory bank $\mathcal{M}$.
    At test time, we select the relevant patch representation via masking (if applicable). The distances of those to the nominal representations in $\mathcal{M}$ give rise to an anomaly map and the corresponding anomaly score $s(\mathbf{x}_\mathrm{test})$ using the aggregation statistic $q$. For both, masking and feature extraction, we utilize DINOv2 ($f$).
    Further examples for other categories are depicted on the right (and in \Cref{fig:MVTec_examples,fig:MVTec_examples2,fig:VisA_examples,fig:VisA_examples2} in \Cref{App:DetailedResults}).}
    \label{fig:teaser}
\end{figure*}

Anomaly detection (AD) in machine learning attempts to identify instances that deviate substantially from the nominal data distribution $p_\mathrm{norm}(x)$.
Anomalies, therefore, raise suspicion of being `generated by a different mechanism' \cite{hawkins1980identification}---often indicating critical, rare, or unforeseen events.
The ability to reliably distinguish anomalies from normal samples is highly valuable across various domains, including security~\cite{siddiqui2019detecting}, healthcare~\cite{medical_ad_survey, medical_ad}, and industrial inspection.
In this work, we focus on the latter, where fully automated systems necessitate the ability to detect defective or missing parts to prevent malfunctions in downstream products, raise alerts for potential hazards, or analyze these to optimize production lines. 
See the right-hand side of \Cref{fig:teaser} for anomalous samples in this context.

AD for industrial images has gained tremendous interest over the last couple of years.
The close-to-optimal results on benchmark data make it seem that the problem of anomaly detection is essentially solved.
For instance, Mousakhan et al.~\cite{mousakhan_2023} report $99.8\%$ and $98.9\%$ AUROC on the popular benchmarks MVTec-AD~\cite{bergmann2021mvtec} and VisA~\cite{zou2022visa}, respectively. 
The most popular AD techniques use the training data to train an anomaly classifier \cite{ruff_deepsad}, or a generative model coupled with reconstruction-based \cite{wolleb2022diffusion, diffusion_models_anomaly, mousakhan_2023}, or likelihood-based \cite{cross_scale_flow, Dae2023sanflow} anomaly scoring.  
However, these approaches operate within the full-shot setting, meaning they rely on access to a sufficiently large amount of training data.
Given the challenges associated with dataset acquisition, the attractivity of a fast and easy-to-deploy methodology, and the requirement to rapidly adapt to covariate shifts in the nominal data distribution~\cite{li2024zero}, there is an increasing interest in few-shot and zero-shot anomaly detection.

Few-shot techniques, however, heavily rely on meaningful features, or as \cite{reiss2022anomaly} frame it: `anomaly detection requires better representations'.
Such better representations are now available with the increasing availability and capabilities of foundation models, i.e., large-scale models trained on massive datasets in unsupervised/self-supervised fashion \cite{clip, caron2021emerging, oquab2024dino2}. 
The performance of few-shot anomaly detection techniques has already been boosted by the use of foundation models, mostly by multimodal approaches incorporating language and vision\cite{jeong2023winclip,cao2023segment,zhou2024anomalyclip,kwak2024few}.

Here, we propose to focus on a vision-only approach, in contrast to such multimodal techniques.
This perspective is motivated by the observation that few-shot anomaly detection is feasible for human annotators based on visual features only, and does not require additional textual description of the given object or the expected types of anomalies (which are typically not known a priori).

Our approach, termed \method, follows the well-established AD framework of \textit{patch-level deep nearest neighbor} 
\cite{yi2020patch,roth2022towards},
and leverages DINOv2 \cite{oquab2024dino2} as a backbone.
We carefully design a suitable preprocessing pipeline for the few-shot scenario, using the zero-shot segmentation abilities of DINOv2 (which alleviates the additional overhead of another segmentation model).
At test time, anomalous samples are detected based on the high distances between their patch representations and the closest counterparts in the nominal memory bank $\mathcal{M}$.

Due to its simplicity, \method{} can be deployed in industrial contexts very easily---in strong contrast to more complex approaches such as \cite{chen2023zero} or \cite{kwak2024few}.
Yet, the proposed method achieves new state-of-the-art performance on anomaly detection in the few-shot regime on MVTec-AD \cite{bergmann2021mvtec} and outperforms all but one competing method on VisA \cite{zou2022visa}.

The structure of the paper is as follows:
\Cref{sec:Related_Work} reviews relevant prior studies and clarifies the distinctions between the settings addressed by zero- and few-shot, and batched zero-shot techniques.
\Cref{sec:Method} introduces our proposed method, \method{}.
An extension of this method to the batched zero-shot scenario is detailed in \Cref{App:Batched-0-Shot}.
\Cref{sec:Experiments} presents the experimental outcomes. Additional results and an ablation study are provided in \Cref{App:DetailedResults,App:Ablation}, respectively.
We address identified failure cases of \method{} in \Cref{App:Limitations}.
The code to reproduce the experiments is available at \href{https://github.com/dammsi/AnomalyDINO}{https://github.com/dammsi/AnomalyDINO}.

\paragraph{Contributions}
\begin{itemize}
    \item We propose \method, a simple and training-free yet highly effective patch-based technique for visual anomaly detection.
    Our method builds upon the high-quality feature representation extracted by DINOv2.
    \item An extensive analysis demonstrates the efficiency and effectiveness of the proposed approach, outperforming other multimodal few-shot techniques in terms of performance \textit{and} inference speed.
    Specifically, \method{} achieves state-of-the-art results for few-shot anomaly detection on MVTec-AD, e.g., pushing the one-shot detection from an AUROC of 93.1\% to 96.6\% (thereby halving the gap between the few- and full-shot setting).
    Moreover, our results on VisA are not only competitive with other few-shot methods but also establish a new state-of-the-art for all training-free few-shot anomaly detectors, achieving the best localization performance across all methods.
\end{itemize}

\section{Related Work}
\label{sec:Related_Work}

\paragraph{Foundation Models for Vision}

Multimodal foundation models have emerged as powerful tools for a wide range of tasks, see e.g., \cite{bommasani2022opportunitiesFM,li2023multimodal,SimCLR,he2022masked,clip,liu2023grounding,openai2024gpt4}.
Most relevant to visual AD are multimodal approaches based on CLIP \cite{clip} or recent LLMs \cite{openai2024gpt4}, but also vision-only approaches like DINO \cite{caron2021emerging,oquab2024dino2}.
CLIP \cite{clip} learns visual concepts from natural language descriptions by training on a dataset of images paired with textual annotations. The model uses a contrastive learning objective that aligns the embeddings from image and text encoders, optimizing the similarity between corresponding image-text pairs. This common feature space for vision and language can be utilized for several downstream tasks, such as zero-shot image classification by assessing similarities to a set of class-specific prompts.
DINO \cite{caron2021emerging,oquab2024dino2} 
leverages a self-supervised student-teacher framework based on vision transformers \cite{dosovitskiy2021an}. It employs a multi-view strategy to train Vision Transformers  (ViT) \cite{vit} to predict softened teacher outputs, thereby learning robust and high-quality features for downstream tasks. DINOv2 \cite{oquab2024dino2} combines ideas from DINO with patch-level reconstruction techniques \cite{zhou2022image} and scales to larger architectures and datasets. The features extracted by DINO are well-suited for anomaly detection as they incorporate both local and global information, are robust to multiple views and crops, and benefit from large-scale pre-training.
GroundingDINO \cite{liu2023grounding}, builds upon the DINO framework and focuses on improving the alignment of textual and visual information, enhancing the model's performance in tasks requiring detailed object localization and multimodal understanding.

\paragraph{Anomaly Detection}

Given a predefined notion of normality, the anomaly detection task is to detect test samples that deviate from this concept~\cite{yang2021generalized, ruff2021unifying}.
In this work, we focus on \textit{low-level sensory anomalies of industrial image data}, i.e., we do not target the detection of semantic anomalies but of low-level features such as scratches of images of industrial products (see e.g., \Cref{fig:teaser}). 
Several works tackled this task by either training an anomaly classifier \cite{ruff_deepsad} or a generative model, which allows for reconstruction-based or likelihood-based AD \cite{wolleb2022diffusion, diffusion_models_anomaly, zhang2023diffusionad, Dae2023sanflow, mousakhan_2023}.   

A well-established approach is that of \textit{deep nearest neighbor} AD, where test instances are scored according to the distance to their nearest neighbors (in feature space) from the memory bank $\mathcal{M}$, containing features of nominal instances.
This approach dates back to (at least) 2002 \cite{eskin2002geometric}, but got later popularized first on image-level \cite{bergmann_dn2}  and then by the seminal works on patch-level:
Patch support vector data description (SVDD)  \cite{yi2020patch} trains a patch-based feature extractor (with a mixture of DeepSVDD \cite{deepSVDD} and a self-supervised loss), and PatchCore  \cite{roth2022towards} utilizes pre-trained classification models on Imagenet for patch-level feature extraction.
Various further approaches build on this simple, but effective AD approach  \cite{spade,xie2023graphcore,Li2024MuSc}. 
Besides kNN matching, other `classical' AD methods like the Mahalanobis distance are employed \cite{padim}. Evidently, the performance of these approaches crucially depends on the feature extractor $f$.
Classical choices are ResNets~\cite{resnet}, Wide ResNets~\cite{wide_resnets}, and Vision Transformers~\cite{vit}, mostly pre-trained on a supervised task.
Another line of work builds upon the success of pre-trained language-vision models in zero-shot classification. 
The underlying idea consists of two steps. First, these approaches define sets of prompts describing nominal samples and anomalies.
Second, the corresponding textual embeddings are compared against the image embeddings~\cite{jeong2023winclip, chen2023zero, kwak2024few, zhou2024anomalyclip}.
Images whose visual embedding is close to the textual embedding of a prompt associated with an anomaly are classified as anomalous. 
However, these methods either require significant prompt engineering (e.g., \cite{chen2023zero} use a total of $35 \times 7$ different prompts for describing normal samples) or fine-tuning of the prompt(-embeddings).
Lastly, another type of few-shot anomaly detection builds upon the success of multimodal chatbots. These methods require more elaborate prompting and techniques for interpreting textual outputs~\cite{xu2024customizing}.   
Since these methods do not require a memory bank, they are capable of performing zero-shot anomaly detection.

\begin{table}[!b]
    \centering
    \caption{\textbf{Taxonomy of recent few- and zero-shot anomaly detection methods}. The $\dagger$ indicates approaches that were introduced as full-shot detectors but then considered as few-shot detectors in later works, see e.g.,~\cite{roth2022towards}.}
    \resizebox{\linewidth}{!}{
    \begin{tabular}{@{}lcccc@{}}
        \toprule
         Method &  Setting & Training Type & Modes \\
         \midrule 
         DN2~\cite{bergmann_dn2} & Few-Shot & Training-Free & Vision \\ 
         SPADE~\cite{spade} \hspace{-1cm}  & \phantom{$^\dagger$}Few-Shot$^\dagger$ & Training-Free & Vision \\ 
         PaDiM~\cite{padim} &  \phantom{$^\dagger$}Few-Shot$^\dagger$ & Fine-Tuning  & Vision \\ PatchCore~\cite{roth2022towards}\hspace{-1cm}& Few-Shot  & Training-Free & Vision \\ 
         PatchCore--opt~\cite{santos2023optimizing}  & Few-Shot & Training-Free & Vision \\
         MuSc~\cite{Li2024MuSc}  & Batched Zero-Shot & Training-Free & Vision \\ 
        GraphCore~\cite{xie2023graphcore}  & Few-Shot & Training (GNN) & Vision \\ 
         WinCLIP~\cite{jeong2023winclip} & Zero-/Few-Shot & Training-Free & Vision + Language \\ 
         APRIL-GAN~\cite{chen2023zero} & Zero-/Few-Shot & Meta-Training & Vision + Language  \\ 
         ADP~\cite{kwak2024few} &Few-Shot & Fine-Tuning &  Vision + Language \\ 
         AnomalyCLIP~\cite{zhou2024anomalyclip} \hspace{-1cm}&Zero-Shot & Meta-Training &  Vision + Language \\        
         GPT4-V~\cite{xu2024customizing}
         &    Zero-Shot & Training-Free & Vision + Language\\  
         ACR~\cite{li2024zero}&   Batched Zero-Shot & Meta-Training & / \\ 
         \midrule 
         \method{} (Ours) &   Few-Shot & Training-Free & Vision \\
         \bottomrule
    \end{tabular}}
    \label{tab:taxonomy}
\end{table}

\paragraph{Categorization of Few-/Zero-Shot Anomaly Detectors}
Previous works consider different AD setups, which complicates their evaluation and comparison.
To remedy this, we provide a taxonomy of recent few- and zero-shot AD based on the particular `shot'-setting,  the training requirements, and the modes covered by the underlying models.
We categorize three `shot'-settings: zero-shot, few-shot, and \textit{batched zero-shot}.
Zero- and few-shot settings are characterized by the number of nominal training samples a method can process, before making predictions on the test samples.
In batched zero-shot, inference is not performed sample-wise but based on a whole batch of test samples, usually the full test set. For instance, the method proposed in \cite{Li2024MuSc} benefits from the fact that a significant majority of pixels correspond to normal pixels, which motivates the strategy of matching patches across a batch of images.
Another work that considers this setting \cite{li2024zero}, deploys a parameter-free anomaly detector based on the effect of batch normalization. 
We split the training requirements into the categories `Training-Free', `Fine-Tuning', and `Meta-Training'. 
`Training-Free' approaches do not require any training, while `Fine-Tuning' methods use the few accessible samples to modify the underlying model. In contrast, `Meta-Training' is associated with training the model on a dataset related to the test data.
For example,~\cite{li2024zero} train their model on MVTec-AD containing all classes except the class they test against. \cite{zhou2024anomalyclip} and~\cite{chen2023zero} train their model on VisA when evaluating the test performance on MVTec-AD and vice versa. 
Finally, we differentiate the leveraged models, which are either vision models (such as pre-trained ViT) or language-vision models (such as CLIP).
We provide a detailed summary in Table~\ref{tab:taxonomy}.

\section{Matching Patch Representations for Visual Anomaly Detection}
\label{sec:Method}

This section introduces \method{}, which leverages DINOv2 to extract meaningful patch-level features. 
We build upon the well-established deep nearest neighbor approaches \cite{eskin2002geometric,yi2020patch,spade,roth2022towards,xie2023graphcore,bergmann_dn2,Li2024MuSc}, i.e., we first gather relevant patch representations of nominal patches in a memory bank $\mathcal{M}$.
Then, for each test patch, we compute its distance to the nearest nominal patch in $\mathcal{M}$.
A suitable aggregation of the patch-based distances gives anomaly scores on image-level.

Our work differs from previous deep nearest neighbor approaches \cite{roth2022towards,yi2020patch} by tailoring the memory bank concept to the few-shot regime, specifically utilizing the strong features from DINOv2: We design a pipeline that incorporates \textit{zero-shot masking} and \textit{augmentations}, and we propose a \textit{more robust aggregation statistic}.
Importantly, we identify DINOv2 as the ideal backbone for our scenario due to its strong patch-level features and masking ability. In addition, this simplifies the deep nearest neighbor framework by reducing the complexity of the feature engineering stage
(e.g., by making it unnecessary to think of which representations/layers to use and how to aggregate them), and \textit{increasing the flexibility} w.r.t.~input resolution (that can be chosen to be any multiple of $14$).
The proposed method is described in detail in the following subsections.

\subsection{Anomaly Detection via Patch (Dis-) Similarities}
Let us briefly review the idea of patch-level deep nearest neighbor AD.
We assume to have access to a suitable feature extractor $f: \mathcal{X} \rightarrow \mathcal{F}^n$ that maps each image $\mathbf{x} \in \mathcal{X}$ to a tuple of patch-features 
$f(\mathbf{x}) = (\mathbf{p}_1, \dots, \mathbf{p}_n)$, where $\mathcal{X}$ denotes some space of images and $\mathcal{F}$ the feature space. 
Note that $n$ depends on the image resolution and the patch size (in the case of DINOv2, a patch is $14\times 14$ pixels).
Given $k\ge1$ nominal reference sample(s) $X_\text{ref}:= \{\mathbf{x}^{(i)} \mid \, i \in [k]\}$ (with shorthand $[k]:=\{1,\dots,k\}$), we collect the nominal patch features and store them in a memory bank
\begin{equation}
    \label{eq:memory_bank}
    \begin{split}
    \mathcal{M}:= \hspace{-2mm} \bigcup_{\mathbf{x}^{(i)} \in X_{\operatorname{ref}}}  \hspace{-2mm}  \bigl\{ \mathbf{p}_j^{(i)} \mid f(\mathbf{x}^{(i)}) = (\mathbf{p}_1^{(i)}, \dots, \mathbf{p}_n^{(i)}),\, j \in [n]\bigr\}  \enspace .
    \end{split}
    \raisetag{2.1ex}
\end{equation}
To score a test sample $\mathbf{x}_\mathrm{test}$, we collect the extracted patch representations $f(\mathbf{x}_\mathrm{test})= (\mathbf{p}_1, \dots, \mathbf{p}_n)$ and then check how well they comply with $\mathcal{M}$. 
To do so, we leverage a nearest neighbor approach to find the distance to the closest reference patch for a given test patch $\mathbf{p} \in \mathcal{F}$
\begin{equation}
\label{eq:NN-distance}
    d_{\operatorname{NN}}(\mathbf{p}
    ; \mathcal{M}) := \min_{\mathbf{p}_{\operatorname{ref}} \in \mathcal{M}} d(\mathbf{p}
    ,\, \mathbf{p}_{\operatorname{ref}})  
\end{equation}
for some distance metric $d$.
In our experiments, we set $d$ as the cosine distance, that is, 
\begin{equation}
\label{eq:cosine_distance}
    d(\mathbf{x}, \mathbf{y}) := 1 - \frac{\langle \mathbf{x},\;  \mathbf{y}\rangle} {\Vert \mathbf{x} \Vert \, \Vert \mathbf{y} \Vert }
    \enspace. 
\end{equation}

The image-level score  $s(\mathbf{x}_\mathrm{test})$ is given by aggregating the patch distances via a suitable statistic $q$
\begin{equation}
    s(\mathbf{x}_\mathrm{test}) := q\big(
    \{
    d_{\operatorname{NN}}(\mathbf{p}_1; \mathcal{M}), 
    \dots, 
    d_{\operatorname{NN}}(\mathbf{p}_n; \mathcal{M})
    \}
    \big) \enspace . 
    \label{eq:meantop1p}
\end{equation}

Throughout this paper, we define $q$ as the average distance of the 1\% most anomalous patches, i.e, \mbox{$q(\mathcal{D}):= \mathrm{mean}(H_{0.01}(\mathcal{D}))$} with $H_{0.01}(\mathcal{D})$ containing the 1\% highest values in the set $\mathcal{D}$. 
The statistic $q$ can be understood as an empirical estimate of the tail value at risk for the 99\% quantile \cite{mcneil1999extreme} and turns out to be suitable for a wide range of settings because it balances two desirable properties:
First, we want $s(\mathbf{x}_\mathrm{test})$ to depend on the highest patch distances, as they may provide the strongest anomaly signal. Similarly, we want a certain degree of robustness against a singular high patch distance (in particular in few-shot scenarios where $\mathcal{M}$ is sparsely populated).
However, one could also replace $q$ with a different statistic to cater to special cases.
If anomalies are expected to cover larger parts of the image, for example, a high percentile of the patch distances might be a suitable choice. 
If, on the contrary, anomalies may occur only locally such that only very few patches/or a single patch might be affected, the score $q$ should be sensitive to the highest patch distances.
Frequently, the maximum pixel-wise anomaly score is considered.
Such an anomaly score on pixel-level is usually obtained by upsampling to full image resolution and applying some smoothing operation.

Following \cite{roth2022towards}, we utilize bilinear upsampling and Gaussian smoothing ($\sigma = 4.0$) to turn the patch distances into pixel-level anomaly scores for localization of potential defects.
Examples of the resulting anomaly maps are visualized in \Cref{fig:teaser} and \Cref{App:DetailedResults} (\Cref{fig:MVTec_examples,fig:MVTec_examples2,fig:VisA_examples,fig:VisA_examples2}).

We extend \method{} also to the batched zero-shot scenario, see \Cref{App:Batched-0-Shot} (\Cref{fig:examples_batched-0-shot}).

\subsection{Enriching the Memory Bank \& Filtering Relevant Patches}

In the few-shot anomaly detection setting, the primary challenge is to effectively capture the concept of normality from the limited set of nominal samples. A useful strategy involves applying simple augmentations, like rotations (following the insights in \cite{xie2023graphcore}), to enhance the variability of nominal patch features. This is essential because the variability at test time is likely to be significantly greater than in the reference data, $X_\mathrm{ref}$. In contrast, full-shot methods aim to reduce the size of $\mathcal{M}$ to reduce inference time, e.g., \cite{roth2022towards}.

To avoid irrelevant areas of the test image from leading to falsely high anomaly scores, we propose masking the object of interest. This approach mitigates the risk of false positives, particularly in the few-shot regime where the limited reference samples may not adequately capture the natural variations in the background, as exemplified in \Cref{fig:masking-background-noise}, \Cref{App:Ablation}.
It is important to note that the appropriate preprocessing technique—whether to apply masking and/or augmentations—should depend on the specific characteristics of the object(s) of interest. For a more detailed discussion on the challenges and considerations in designing an effective preprocessing pipeline, see \Cref{App:Ablation-Preprocessing}.

\paragraph{Masking}

\begin{figure}[!htb]
\centering 
    \includegraphics[width=0.49\linewidth]{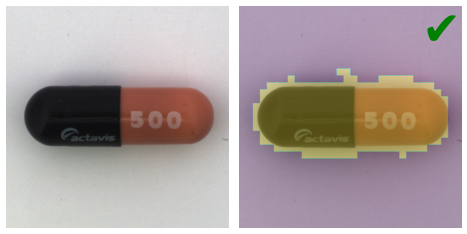} 
    \includegraphics[width=0.49\linewidth]{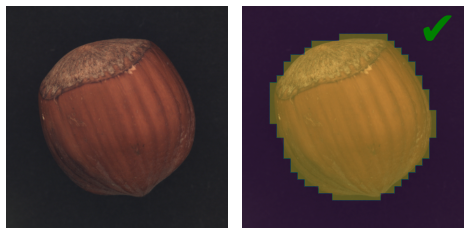} 
    \includegraphics[width=0.49\linewidth]{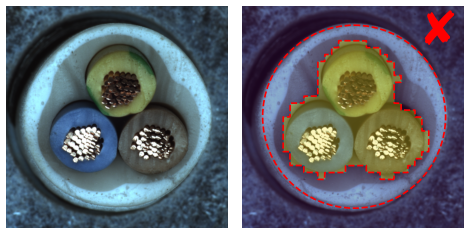}
    \includegraphics[width=0.49\linewidth]{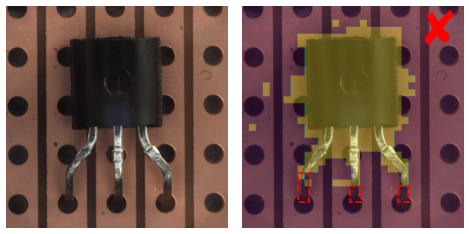}
    \caption{\textbf{Masking test on MVTec-AD}. For `Capsule' and `Hazelnut' the masking works successfully (top row), while for `Cable' and `Transistor' (bottom row) some areas are incorrectly predicted as background that should belong to the object of interest (highlighted in red). See App.~\ref{App:Ablation-Preprocessing} for the outcomes per object.}  
    \label{fig:Preprocessing_Masking_Test}
    \vspace{-3mm}
\end{figure}

Masking, i.e., delineating the primary object(s) in an image from its background, can help to reduce false-positive predictions, thereby improving the robustness in the low-data regime. 
To minimize the overhead of the proposed pipeline, we utilize DINOv2 also for masking.
This is achieved by thresholding the first PCA component of the patch features \cite{oquab2024dino2}.
We observe empirically that this sometimes produces erroneous masks.
Frequently, such failure cases occur for close-up shots, where the objects of interest account for $\gtrsim 50\%$ of patches.
To address this issue, we check if the PCA-based mask accurately captures the object in the first reference sample and apply the mask accordingly,
which gives rise to the `masking test' in \Cref{fig:Preprocessing_Masking_Test}. 
This test is performed \textit{only once} per object as the procedure yields very consistent outputs.
In addition, we utilize dilation and morphological closing to eliminate small holes and gaps within the predicted masks.
See \Cref{fig:MaskingExamplesMVTec,fig:MaskingExamplesVisA} for examples of this masking procedure, \Cref{tab:Preprocessing_Default} for the outcomes of the masking test per object, and \Cref{fig:masking-background-noise} for a visualization of the benefits of masking in the presence of background noise (all in \Cref{App:Ablation}).
In general, we do not mask textures (e.g., `Wood' or `Tile' in MVTec-AD).

\paragraph{Rotation}
Rotating the reference sample may improve the detection performance by better resembling the variations within the concept of normality captured in $\mathcal{M}$. Consider, e.g., the `Screw' in MVTec-AD as depicted in \Cref{fig:teaser} for which rotation-invariant features are desirable.
On the other hand, rotation can also be detrimental in cases where rotations of (parts of) the object of interest can be considered anomalies themselves (see \Cref{fig:Transistor_Rotation} for an example).

We consider two different settings.
In the `agnostic' case, which we focus on in the main paper, we always augment the reference sample with rotations. 
We also consider the case where we know about the samples' potential rotations (`informed').
The `informed' case is sensible as the data collection process can usually be controlled in an industrial/medical setting (or test images can be aligned), and may lead to lower inference time. 
See \Cref{App:Ablation-Preprocessing} for the effect of masking and rotation, and a comparison of `informed' vs. `agnostic'.

\section{Experiments}
\label{sec:Experiments}

\paragraph{\method{} -- Defaults}
Using DINOv2\footnote{Code/weights of \href{https://github.com/facebookresearch/dinov2}{DINOv2} \cite{oquab2024dino2} are available under Apache 2.0 license.} as backbone allows us to choose from different distillation sizes, which range from small (ViT-S, $21\times10^6$ parameters) to giant (ViT-G, $1.1\times10^9$ parameters). 
To prioritize low latency, we take the smallest model as our default (and denote our default pipeline \method{}-S, accordingly) and evaluate two input resolutions, 448 and 672 pixels (smaller edge).
As discussed above, we utilize the `agnostic' preprocessing by default (see \Cref{App:Ablation}).

\paragraph{Datasets} For the experiments, we consider MVTec-AD\footnote{The \href{https://www.mvtec.com/company/research/datasets/mvtec-ad}{MVTec-AD dataset} \cite{bergmann2021mvtec} is available under CC BY-NC-SA 4.0.} 
and VisA,\footnote{The \href{https://github.com/amazon-science/spot-diff}{VisA dataset} \cite{zou2022visa} is released under the CC BY 4.0 license.} two datasets with high-resolution images for industrial anomaly detection.
MVTec-AD consists of fifteen categories, depicting single objects or textures, and up to eight anomaly types per category. 
VisA has twelve categories and comes with diverse types of anomalies condensed into one anomaly class `bad'. Some categories in VisA can be considered more challenging as multiple objects and complex structures are present.

\paragraph{Evaluation Metrics}
We assess the image-level detection and the pixel-level segmentation performance with three metrics each, following \cite{jeong2023winclip, chen2023zero, Li2024MuSc}.
For evaluating the detection performance, we measure the area under the receiver-operator curve (AUROC), the F1-score at the optimal threshold (F1-max), and the average precision (AP) using the respective image-level anomaly scores. We quantify the anomaly segmentation performance using the \mbox{AUROC}, F1-max, and the per-region overlap (PRO, \cite{bergmann2021mvtec}) of the segmentation using the pixel-wise anomaly scores.
Note that due to a high imbalance of nominal and anomalous pixels---for MVTec-AD we have 97.26\% nominal pixel, for VisA even 99.45\% \cite{Li2024MuSc}---it is not recommended to assess performance solely on segmentation \mbox{AUROC}~\cite{rafiei2023pixel}.
We repeat each experiment three times and report mean and standard deviation.\footnote{The randomness stems from the choice of reference samples $X_{\mathrm{ref}}$.
For straightforward reproducibility, we simply set $X_{\mathrm{ref}}$ to the first $k$, second $k$ and third $k$ train samples for the three different runs, respectively.}

\paragraph{Baselines}
We compare \method{} with a range of modern zero- and few-shot AD models, e.g., SPADE~\cite{spade}, PaDiM~\cite{padim}, PatchCore~\cite{roth2022towards}, 
WinCLIP+~\cite{jeong2023winclip}, and APRIL-GAN~\cite{chen2023zero}. 
It is important to note that ACR~\cite{li2024zero} and MuSc~\cite{Li2024MuSc} consider the batched zero-shot setting (see Table~\ref{tab:taxonomy}), thus, covering a different setting than \method{}. We adapt \method{} to this setting in \Cref{App:Batched-0-Shot}.
Moreover, APRIL-GAN and AnomalyCLIP, require training on a related dataset, which is in contrast to our proposed training-free method. 
In \Cref{tab:results_MVTec,tab:results_VisA}, results reported from \cite{jeong2023winclip} are indicated by $^\dagger$,
the result of the WinCLIP re-implementation \cite{kwak2024few} (where J. Jeong is also coauthor) by {$^*$}. All other results are taken from the original publication.
For ADP \cite{kwak2024few} we report the (usually slightly better) variant ADP$_\ell$ (with access to the class label). 
For GraphCore no standard deviations are reported \cite{xie2023graphcore}.

\subsection{Few-Shot Anomaly Detection and Anomaly Segmentation}

\paragraph{Results}

We summarize the results for few-shot anomaly detection on MVTec-AD and VisA in Table~\ref{tab:results_MVTec} and Table~\ref{tab:results_VisA}, respectively. 
Regarding MVTec-AD, our method achieves state-of-the-art $k$-shot detection performance across all $k \in \{1, 2, 4, 8, 16\}$ for every reported metric, outperforming approaches that require additional training data sets (such as ADP and APRIL-GAN). The method also demonstrates superior anomaly localization, with results showing that while detection performance remains comparable across different resolutions (448 vs. 672), a higher resolution enhances localization performance. Furthermore, we observe clear improvements as the number of samples increases.

\noindent\setlength\extrarowheight{1.25pt}

\begin{table}[!bht]
\centering 
\caption{\textbf{Anomaly detection on MVTec-AD}. Quantitative results for detection (image-level) and segmentation (pixel-level). 
For each shot, we highlight the \textbf{best result} in bold, the results from the {\ul{second best}} method as underlined, and the \colorbox{\hcolor}{best training-free} result by a gray box (see also \Cref{tab:taxonomy}). All results in \%.}
\vspace{-2mm}
\resizebox{\linewidth}{!}{

\begin{tabular}{@{}lccccccc}
\toprule
\multirow{2}{*}{Method}
& \multicolumn{3}{c}{Classification}                                                                       &                      & \multicolumn{3}{c}{Segmentation}                                                                        \\ \cmidrule(l){2-8} 
                        & \multicolumn{1}{c}{AUROC}         & \multicolumn{1}{c}{F1-max}        & \multicolumn{1}{c}{AP}           & \multicolumn{1}{c}{} & \multicolumn{1}{c}{AUROC}        & \multicolumn{1}{c}{F1-max}       & \multicolumn{1}{c}{PRO}          \\ \hline 

\multicolumn{8}{c}{ \textbf{1-shot}}                                                                                                                                                                                                                                 \\ \hline
SPADE$^\dagger$         & 81.0\std{2.0}             & 90.3\std{0.8}              & 90.6\std{0.8}           &  & 91.2\std{0.4}             & 42.4\std{1.0}              & 83.9\std{0.7}           \\
PatchCore$^\dagger$     & 83.4\std{3.0}             & 90.5\std{1.5}              & 92.2\std{1.5}           &  & 92.0\std{1.0}             & 50.4\std{2.1}              & 79.7\std{2.0}           \\
GraphCore               & 89.9\std{\printtowidth{1.0}{/}}             &/                         &/                      &  & {\ul 95.6\std{\printtowidth{1.0}{/}}}                       &/                         &/                      \\
 
WinCLIP+                & {\ul 93.1\std{2.0}}       & {\ul 93.7\std{1.1}}        & {\ul 96.5\std{0.9}}     &  & {95.2\std{0.5}}       & {\ul 55.9\std{2.7}}        & 87.1\std{1.2}           \\
APRIL-GAN               & 92.0\std{0.3}             & 92.4\std{0.2}              & 95.8\std{0.2}           &  & 95.1\std{0.1}             & 54.2\std{0.0}              & {\ul 90.6\std{0.2}}     \\ 
\method{}-S \scriptsize{(448)}                    & 96.5\std{0.4} & \textbf{96.0\std{0.2}} \cellcolor{\hcolor} & 98.1\std{0.3}  & & 96.3\std{0.1} & 57.9\std{0.8} & 91.7\std{0.1} \\
\method{}-S \scriptsize{(672)}                    & \textbf{96.6\std{0.4}} \cellcolor{\hcolor} & 95.8\std{0.5} & \textbf{98.2\std{0.2}} \cellcolor{\hcolor}  & & \textbf{96.8\std{0.1}} \cellcolor{\hcolor}  & \textbf{60.2\std{1.1}} \cellcolor{\hcolor}  & \textbf{92.7\std{0.1}} \cellcolor{\hcolor}  \\ \hline
\multicolumn{8}{c}{\textbf{2-shot}} \\ \hline

SPADE$^\dagger$         & 82.9\std{2.6}             & 91.1\std{1.0}              & 91.7\std{1.2}           &  & 92.0\std{0.3}             & 44.5\std{1.0}              & 85.7\std{0.7}           \\
PatchCore$^\dagger$     & 86.3\std{3.3}             & 92.0\std{1.5}              & 93.8\std{1.7}           &  & 93.3\std{0.6}             & 53.0\std{1.7}              & 82.3\std{1.3}           \\
GraphCore               & 91.9\std{\printtowidth{1.0}{/}}             &/                         &/                      &  & {\ul 96.9\std{\printtowidth{1.0}{/}}}                       &/                         &/                      \\
WinCLIP+                & 94.4\std{1.3}             & {\ul 94.4\std{0.8}}        & {\ul 97.0\std{0.7}}     &  & {96.0\std{0.3}}       & {\ul 58.4\std{1.7}}        & 88.4\std{0.9}           \\
ADP$_\ell$              & {\ul 95.4\std{0.9}}       &/                         &/                      &  &/                        &/                         &/                      \\
APRIL-GAN               & 92.4\std{0.3}             & 92.6\std{0.1}              & 96.0\std{0.2}           &  & 95.5\std{0.0}             & 55.9\std{0.5}              & {\ul 91.3\std{0.1}}     \\ 
\method{}-S \scriptsize{(448)}    & 96.7\std{0.8} & \textbf{96.5\std{0.4}}  \cellcolor{\hcolor} & 98.1\std{0.7} &  & 96.5\std{0.2} & 58.5\std{0.5} & 92.0\std{0.2} \\
\method{}-S \scriptsize{(672)}    & \textbf{96.9\std{0.7}} \cellcolor{\hcolor}  & 96.1\std{0.3} & \textbf{98.2\std{0.5}} \cellcolor{\hcolor}  & &  \textbf{97.0\std{0.2}} \cellcolor{\hcolor}  & \textbf{61.0\std{0.5}} \cellcolor{\hcolor}  & \textbf{93.1\std{0.2}} \cellcolor{\hcolor}  \\ \hline
\multicolumn{8}{c}{\textbf{4-shot}}         \\ \hline                                                                                                                                                                                                                                     
SPADE$^\dagger$         & 84.8\std{2.5}             & 91.5\std{0.9}              & 92.5\std{1.2}           &  & 92.7\std{0.3}             & 46.2\std{1.3}              & 87.0\std{0.5}           \\
PatchCore$^\dagger$     & 88.8\std{2.6}             & 92.6\std{1.6}              & 94.5\std{1.5}           &  & 94.3\std{0.5}             & 55.0\std{1.9}              & 84.3\std{1.6}           \\
GraphCore          & 92.9\std{\printtowidth{1.0}{/}}             &/                         &/                      &  & \textbf{97.4\std{\printtowidth{1.0}{/}}}                       &/                         &/                      \\
  
WinCLIP+                & 95.2\std{1.3}             & {\ul 94.7\std{0.8}}        & {\ul 97.3\std{0.6}}     &  & {96.2\std{0.3}}       & {\ul 59.5\std{1.8}}     & 89.0\std{0.8}           \\
ADP$_\ell$              & {\ul 96.2\std{0.8}}       &/                         &/                      &  &/                        &/                         &/                      \\
APRIL-GAN               & 92.8\std{0.2}             & 92.8\std{0.1}              & 96.3\std{0.1}           &  & 95.9\std{0.0}             & 56.9\std{0.1}              & {\ul 91.8\std{0.1}}     \\ 
\method{}-S \scriptsize{(448)} & 97.6\std{0.1} & \textbf{97.0\std{0.3}} \cellcolor{\hcolor}& 98.4\std{0.3} & & 96.7\std{0.1} & 59.2\std{0.4} & 92.4\std{0.1} \\
\method{}-S \scriptsize{(672)} & \textbf{97.7\std{0.2}} \cellcolor{\hcolor} & 96.6\std{0.0} & \textbf{98.7\std{0.1}} \cellcolor{\hcolor} & & {\ul 97.2\std{0.1}} \cellcolor{\hcolor} & \textbf{61.8\std{0.1}}\cellcolor{\hcolor} & \textbf{93.4\std{0.1}} \cellcolor{\hcolor} \\ \hline
\multicolumn{8}{c}{\textbf{8-shot}}                                                                                                                                                                                                                                          \\ \hline
GraphCore          & 95.9\std{\printtowidth{1.0}{/}}             &/                         &/                      &  & \textbf{97.8\std{\printtowidth{1.0}{/}}}                       &/                         &/                      \\

WinCLIP+                & \phantom{\tiny{$^*$}}94.6\std{0.1}{$^*$}             &/                         &/                      &  &/                        &/                         &/                      \\
ADP$_\ell$              & {\ul 97.0\std{0.2}}       &/                         &/                      &  &/                        &/                         &/                      \\
APRIL-GAN               & 93.1\std{0.2}             & {\ul 93.1\std{0.2}}        & {\ul 96.4\std{0.2}}     &  & {96.2\std{0.1}}       & {\ul 57.7\std{0.2}}       & {\ul 92.4\std{0.2}}     \\ 
\method{}-S \scriptsize{(448)} & 98.0\std{0.1} & \textbf{97.4\std{0.1}} \cellcolor{\hcolor} & 99.0\std{0.2} & & 97.0\std{0.1} & 59.6\std{0.2} & 92.7\std{0.0} \\
\method{}-S \scriptsize{(672)} & \textbf{98.2\std{0.2}} \cellcolor{\hcolor}  & \textbf{97.4\std{0.2}} \cellcolor{\hcolor} & \textbf{99.1\std{0.1}} \cellcolor{\hcolor}  & & {\ul 97.4\std{0.1}}\cellcolor{\hcolor}  & \textbf{62.3\std{0.1}} \cellcolor{\hcolor}  & \textbf{93.8\std{0.1}} \cellcolor{\hcolor}  \\ \hline
\multicolumn{8}{c}{\textbf{16-shot}}      \\ \hline                                                       
WinCLIP+                & \phantom{\tiny{$^*$}}94.8\std{0.1}{$^*$}             &/                         &/                      &  &/                        &/                         &/                      \\
ADP$_\ell$              & {\ul 97.0\std{0.3}  }           &/                         &/                      &  &/                        &/                         &/                      \\
APRIL-GAN               & 93.2\std{0.1}             & {\ul 93.0\std{0.1}}        & {\ul 96.5\std{0.1 }}    &  & {\ul 96.4\std{0.0}}       & {\ul 58.5\std{0.1}}        & {\ul 92.6\std{0.1}}    \\ 
\method{}-S \scriptsize{(448)} & 98.3\std{0.1} & \textbf{97.7\std{0.2}} \cellcolor{\hcolor} & 99.3\std{0.0} & & 97.1\std{0.1} & 60.0\std{0.1} & 92.9\std{0.1} \\
\method{}-S \scriptsize{(672)} & \textbf{98.4\std{0.1}} \cellcolor{\hcolor} & 97.6\std{0.1} & \textbf{99.3\std{0.0}} \cellcolor{\hcolor} & & \textbf{97.5\std{0.0}} \cellcolor{\hcolor} & \textbf{62.7\std{0.1}} \cellcolor{\hcolor} & \textbf{94.0\std{0.1}} \cellcolor{\hcolor} \\ \bottomrule
\end{tabular}}
\label{tab:results_MVTec}
\vspace{-3mm}
\end{table}

\begin{table}[!htb]
\centering
\caption{\textbf{Anomaly detection on VisA}. Quantitative results for detection (image-level) and segmentation (pixel-level).
For each shot, we highlight the \textbf{best result} in bold, the results from the {\ul{second best}} method as underlined, and the \colorbox{\hcolor}{best training-free} result by a gray box (see also \Cref{tab:taxonomy}). All results in \%.}
\vspace{-2mm}
\resizebox{\linewidth}{!}{
\begin{tabular}{@{}lccccccc}
\toprule
\multirow{2}{*}{Method}
& \multicolumn{3}{c}{Classification}                                                                       &                      & \multicolumn{3}{c}{Segmentation}                                                                        \\ \cmidrule(l){2-8} 
                        & \multicolumn{1}{c}{AUROC}         & \multicolumn{1}{c}{F1-max}        & \multicolumn{1}{c}{AP}           & \multicolumn{1}{c}{} & \multicolumn{1}{c}{AUROC}        & \multicolumn{1}{c}{F1-max}       & \multicolumn{1}{c}{PRO}          \\ \hline 
\multicolumn{8}{c}{ \textbf{1-shot}}      \\ \hline
SPADE$^\dagger$         & 79.5\std{4.0}             & 80.7\std{1.9}              & 82.0\std{3.3}          &  & 95.6\std{0.4}             & 35.5\std{2.2}              & 84.1\std{1.6}           \\
PaDiM$^\dagger$         & 62.8\std{5.4}             & 75.3\std{1.2}              & 68.3\std{4.0}          &  & 89.9\std{0.8}             & 17.4\std{1.7}              & 64.3\std{2.4}           \\
PatchCore$^\dagger$     & 79.9\std{2.9}             & 81.7\std{1.6}              & 82.8\std{2.3}          &  & 95.4\std{0.6}             & 38.0\std{1.9}              & 80.5\std{2.5}           \\
WinCLIP+                & 83.8\std{4.0}             & 83.1\std{1.7}              & 85.1\std{4.0}          &  & {\ul 96.4\std{0.4}}    & {\ul 41.3\std{2.3}}              & 85.1\std{2.1}           \\
APRIL-GAN               & \textbf{91.2\std{0.8}}    & \textbf{86.9\std{0.6}}     & \textbf{93.3\std{0.8}} &  & 96.0\std{0.0}             & 38.5\std{0.3}              & {\ul 90.0\std{0.1}}  \\ 
\method{}-S \scriptsize{(448)}               & 85.6\std{1.5} & 83.1\std{1.1} & 86.6\std{1.3} && 97.5\std{0.1} & 41.9\std{0.5} & 90.7\std{0.5} \\
\method{}-S \scriptsize{(672)}               & {\ul 87.4\std{1.2}}  \cellcolor{\hcolor}& {\ul 84.3\std{0.5}} \cellcolor{\hcolor} & {\ul 89.0\std{1.0}}  \cellcolor{\hcolor}&& \textbf{97.8\std{0.1}}  \cellcolor{\hcolor} & \textbf{45.1\std{0.9}} \cellcolor{\hcolor} & \textbf{92.5\std{0.5}} \cellcolor{\hcolor} \\ \hline

\multicolumn{8}{c}{ \textbf{2-shot}}      \\ \hline
SPADE$^\dagger$     & 80.7\std{5.0}             & 81.7\std{2.5}              & 82.3\std{4.3}          &  & 96.2\std{0.4}             & 40.5\std{3.7}              & 85.7\std{1.1}           \\
PaDiM$^\dagger$         & 67.4\std{5.1}             & 75.7\std{1.8}              & 71.6\std{3.8}          &  & 92.0\std{0.7}             & 21.1\std{2.4}              & 70.1\std{2.6}           \\
PatchCore$^\dagger$     & 81.6\std{4.0}             & 82.5\std{1.8}              & 84.8\std{3.2}          &  & 96.1\std{0.5}             & 41.0\std{3.9}              & 82.6\std{2.3}           \\
WinCLIP+                & 84.6\std{2.4}             & 83.0\std{1.4}              & 85.8\std{2.7}          &  & {\ul 96.8\std{0.3}}    & {\ul 43.5\std{3.3}}        & 86.2\std{1.4}           \\
ADP$_\ell$              & 86.9\std{0.9}             &/                         &/                     &  &/                        &/                         &/                      \\
APRIL-GAN               & \textbf{92.2\std{0.3}}    & \textbf{87.7\std{0.3}}     & \textbf{94.2\std{0.3}} &  & 96.2\std{0.0}             & 39.3\std{0.2}              & {\ul 90.1\std{0.1}}  \\ 
\method{}-S \scriptsize{(448)}                   & 88.3\std{1.8} & 84.8\std{1.2} & 89.2\std{1.3} && 97.8\std{0.1} & 44.2\std{0.3} & 91.7\std{0.5} \\
\method{}-S \scriptsize{(672)}                   & {\ul 89.7\std{1.3}} \cellcolor{\hcolor} & {\ul 86.3\std{1.2}} \cellcolor{\hcolor} & {\ul 90.7\std{0.8}} \cellcolor{\hcolor} && \textbf{98.0\std{0.1}} \cellcolor{\hcolor} & \textbf{47.6\std{0.5}} \cellcolor{\hcolor} & \textbf{93.4\std{0.6}} \cellcolor{\hcolor} \\\hline

\multicolumn{8}{c}{ \textbf{4-shot}}      \\ \hline
SPADE$^\dagger$         & 81.7\std{3.4}             & 82.1\std{2.1}              & 83.4\std{2.7}          &  & 96.6\std{0.3}             & 43.6\std{3.6}              & 87.3\std{0.8}           \\
PaDiM$^\dagger$         & 72.8\std{2.9}             & 78.0\std{1.2}              & 75.6\std{2.2}          &  & 93.2\std{0.5}             & 24.6\std{1.8}              & 72.6\std{1.9}           \\
PatchCore$^\dagger$     & 85.3\std{2.1}             & 84.3\std{1.3}              & 87.5\std{2.1}          &  & {96.8\std{0.3}}       & 43.9\std{3.1}              & 84.9\std{1.4}           \\
WinCLIP+                & 87.3\std{1.8}             & 84.2\std{1.6}              & 88.8\std{1.8}          &  & {\ul 97.2\std{0.2}}    & {\ul 47.0\std{3.0}}     & 87.6\std{0.9}     \\
ADP$_\ell$              & 88.4\std{0.4}             &/                         &/                     &  &/                        &/                         &/                      \\
APRIL-GAN               & {\ul \textbf{92.6\std{0.4}}}    & {\ul 88.4\std{0.5}}     & \textbf{94.5\std{0.3}} &  & 96.2\std{0.0}             & 40.0\std{0.1}              & {\ul 90.2\std{0.1}}  \\ 
\method{}-S \scriptsize{(448)}    & 91.3\std{0.8} & 87.5\std{1.0} & 91.8\std{0.7} && 98.0\std{0.0} & 46.1\std{0.3} & 92.5\std{0.2} \\
\method{}-S \scriptsize{(672)}    & {\ul \textbf{92.6\std{0.9}}} \cellcolor{\hcolor} & \textbf{88.8\std{0.9}} \cellcolor{\hcolor} & {\ul 92.9\std{0.7}} \cellcolor{\hcolor} && \textbf{98.2\std{0.0}} \cellcolor{\hcolor} & \textbf{49.4\std{0.3}} \cellcolor{\hcolor} & \textbf{94.1\std{0.1}} \cellcolor{\hcolor} \\ \hline

\multicolumn{8}{c}{ \textbf{8-shot}}      \\ \hline
WinCLIP+                & \phantom{\tiny{$^*$}}85.0\std{0.0}{$^*$}             &/                         &/                     &  &/                        &/                         &/                      \\
ADP$_\ell$              & 89.2\std{0.1}             &/                         &/                     &  &/                        &/                         &/                      \\
APRIL-GAN               & {\ul 93.0\std{0.2}}    & {\ul 88.8\std{0.2}}        & \textbf{94.9\std{0.3}} &  & {\ul 96.3\std{0.0}}       & {\ul 40.2\std{0.1}}        & {\ul 90.2\std{0.0}}  \\ 
\method{}-S \scriptsize{(448)} & 92.6\std{0.1} & 88.6\std{0.2} & 92.9\std{0.2} && 98.2\std{0.0} & 47.6\std{0.5} & 93.3\std{0.2} \\
\method{}-S \scriptsize{(672)} & \textbf{93.8\std{0.3}}  \cellcolor{\hcolor} & \textbf{90.0\std{0.1}} \cellcolor{\hcolor} & {\ul 94.3\std{0.4}} \cellcolor{\hcolor} && \textbf{98.4\std{0.0}} \cellcolor{\hcolor} & \textbf{51.1\std{0.4}} \cellcolor{\hcolor} & \textbf{94.8\std{0.2}}\cellcolor{\hcolor} \\ \hline

\multicolumn{8}{c}{\textbf{16-shot}}      \\ \hline
WinCLIP+                & \phantom{\tiny{$^*$}}85.0\std{0.1}{$^*$}             &/                         &/                     &  &/                        &/                         &/                      \\
ADP$_\ell$              & 90.1\std{0.5}             &/                         &/                     &  &/                        &/                         &/                      \\
APRIL-GAN               & {\ul 93.2\std{0.2}}       & {\ul 89.0\std{0.1}}        & {\ul 95.2\std{0.1}} &  & {\ul 96.3\std{0.0}}       & {\ul 40.6\std{0.1}}        & {\ul 90.2\std{0.1}}  \\ 
\method{}-S \scriptsize{(448)} & 93.8\std{0.1} & 89.9\std{0.3} & 94.2\std{0.3} && 98.3\std{0.0} & 48.6\std{0.3} & 93.8\std{0.2} \\
\method{}-S \scriptsize{(672)} & \textbf{94.8\std{0.2}} \cellcolor{\hcolor} & \textbf{90.9\std{0.2}} \cellcolor{\hcolor} & \textbf{95.3\std{0.3}} \cellcolor{\hcolor} && \textbf{98.5\std{0.0}} \cellcolor{\hcolor} & \textbf{52.5\std{0.5}} \cellcolor{\hcolor} & \textbf{95.3\std{0.2}} \cellcolor{\hcolor} \\ \bottomrule
\end{tabular}}
\label{tab:results_VisA}
\vspace{-4mm}
\end{table}

In terms of anomaly detection in the VisA benchmark (\Cref{tab:results_VisA}), APRIL-GAN demonstrates the strongest performance for \mbox{$k \in \{1,2\}$}. 
Nonetheless, \method{} consistently achieves second-best results for \mbox{$k \in \{1,2\}$}, comparable results in the 4-shot setting, and sets new state-of-the-art for $k \in \{8,16\}$. 
This can be attributed to \method{}'s ability to benefit more from a richer memory bank $\mathcal{M}$ than APRIL-GAN. 
We hypothesize that meta-learning exerts a greater influence on APRIL-GAN (i.e., training on MVTec-AD, when testing on VisA, and vice-versa) compared to learning from the nominal features of the given reference samples. Note also that \method{} outperforms all other training-free approaches.

Comparing the segmentation performance on VisA, Table~\ref{tab:results_VisA} reveals a clear picture: AnomalyDINO consistently shows the strongest localization performance in all metrics considered.
While \method{}-S (448) already demonstrates strong performance, the advantages of using a higher resolution (672) become more evident on the VisA dataset.
We attribute this fact to smaller anomalous regions (for which smaller effective patch sizes are beneficial) and more complex scenes (compared to MVTec-AD).

We have further adapted \method{} to the batched zero-shot setting (see \Cref{App:Batched-0-Shot}).
This adaptation is straightforward and relatively simple, especially compared to MuSc. Notably, we did not employ additional techniques such as `Re-scoring with Constrained Image-level Neighborhood' or `Local
Neighborhood Aggregation with Multiple Degrees' \cite{Li2024MuSc}. 
The results obtained, as detailed in \Cref{tab:results_other_settings}, are therefore quite satisfactory.

We conclude that AnomalyDINO---despite its simplicity---rivals other methods in all settings and even comes out ahead in most of the settings (e.g., significantly reducing the gap between few-shot and full-shot methods for MVTec-AD). Within the class of training-free models, it is the clear winner essentially across the board. These results do not only demonstrate the virtues of AnomalyDINO itself but, more generally, highlight the merits of strong visual features as compared to highly engineered architectures. 
We provide further qualitative results in \cref{App:DetailedResults}, and discuss the limitations (such as detecting semantic anomalies) and specific failure cases of our approach in~\Cref{App:Limitations}.

\begin{table}[!hbt]
\caption{\textbf{Detection results for other settings}. All results are \mbox{AUROC} values (in \%).}
\vspace{-2mm}
    \centering
    \resizebox{0.95\linewidth}{!}{
    \begin{tabular}{@{}llcc@{}}
    \toprule
        Setting & Method      & MVTec-AD       & VisA   \\ \midrule 
        \multirow{3}{*}{0-shot} & WinCLIP     & 91.8  & 78.1 \\
        & AnomalyCLIP \hspace{-1cm} & 91.5         & 82.1 \\
        &APRIL-GAN  \hspace{-4mm} & 86.1 & 78.0 \\ 
        \midrule 
        \multirow{4}{*}{Batched 0-shot\hspace{-2mm}}& ACR & 85.8 &  / \\ 
        & MuSc & 97.8 & 92.8 \\ 
        & \method{}-S \scriptsize{(448)} \hspace{-4mm} & 93.0 & 89.7 \\
        & \method{}-S \scriptsize{(672)} \hspace{-4mm} & 94.2 & 90.7 \\
         \bottomrule
    \end{tabular}}
    \vspace{-3mm}
\label{tab:results_other_settings}
\end{table}

\subsection{Ablation Study}
\begin{figure}[!htb]
    \centering
    \vspace{-3mm}
    \includegraphics[width=1.0\linewidth]{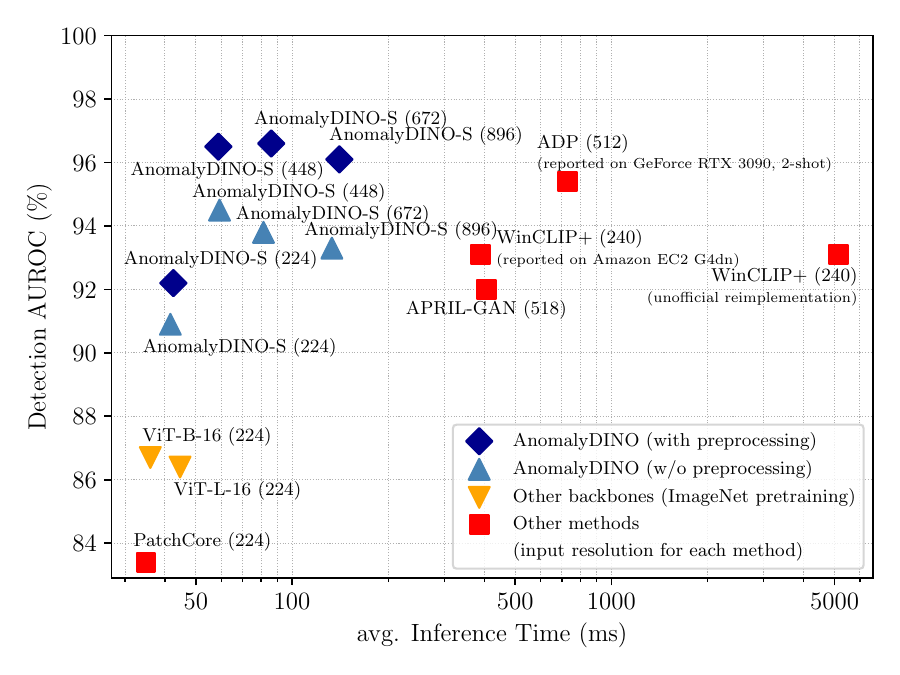}
    \vspace{-6mm}
    \caption{\textbf{Detection AUROC vs. inference time} per sample on MVTec-AD in the 1-shot setting. The input resolution is given in parentheses after the method name. 
    All runtimes are measured on a single NVIDIA A40 if not stated otherwise. (Note that for ADP and WinCLIP+ no official code is available.)}
    \label{fig:detection_vs_inferencetime}
    \vspace{-6mm}
\end{figure}

We conduct additional experiments to assess the effect of specific design choices in our pipeline. 
The full ablation study is provided in \Cref{App:Ablation}, here we briefly summarize the main insights.
\Cref{fig:detection_vs_inferencetime} supports the ablation study by highlighting the two key aspects: performance and runtime.

\paragraph{Inference time}
Comparing the inference times of \method{} with the other approaches in the 1-shot setting (see \Cref{fig:detection_vs_inferencetime}), we observe that \method{} achieves significantly faster inference times than SOTA few-shot competitors (note the logarithmic scale).
For example, \method{}-S takes roughly 60ms to process an image at a resolution of 448.
The only approaches with lower inference time are PatchCore and \method{} with ViT-backbones trained on ImageNet, which however sacrifice performance.
Notably, while augmentations increase the memory bank, and thus, the potential runtime of the nearest neighborhood search, we find that the effect is negligible in practice.\footnote{Our implementation leverages GPU-accelerated neighborhood search~\cite{johnson2019billion}, leading to only slightly increased inference time. With increasing size of $\mathcal{M}$ reduction techniques such as coreset subsampling \cite{roth2022towards} are advisable.} A more detailed runtime analysis is included in \Cref{App:Ablation}, \Cref{tab:runtime}.

\paragraph{Preprocessing}
To study the impact of preprocessing, we compare the resulting detection \mbox{AUROCs} of \method{} with and without preprocessing in~\Cref{fig:detection_vs_inferencetime}. Across four different configurations of \method{}, we observe that by incorporating the proposed preprocessing, the resulting AUROCs increase by approximately $2\%$ without significantly affecting the inference time.
The effect of proper preprocessing seems to increase with higher resolutions.
Additionally, we compare the two different preprocessing settings, `agnostic' (our default) and `informed', and find that the agnostic preprocessing slightly outperforms the informed counterpart, see \Cref{fig:ablation_preprocessing_MVTec,fig:ablation_preprocessing_VisA}. 
Depending on the product category, suitable preprocessing steps lead to substantial improvements. 
And while in principle augmenting the reference sample with rotations may negatively impact the detection performance, we observe this in only very few examples.
For the full discussion, see \Cref{App:Ablation-Preprocessing}.

\paragraph{Aggregation statistic}
We compare the default scoring method, the empirical tail value at risk, to other suitable aggregations in \Cref{App:Ablation-Scoring}. 
The mean of the 1\% highest distances from test patches to $\mathcal{M}$ improves over the standard choice (maximum of the upsampled and smoothed patch distances).

\paragraph{Architecture size/choice}
We also evaluated the proposed framework to Vision Transformers pre-trained on \mbox{ImageNet}, see \Cref{fig:detection_vs_inferencetime} and \cref{app:ablation-backbone}. While these backbones slightly outperform PatchCore, they give weaker features compared to DINOv2 and are incompatible with the proposed masking procedure.
This underlines that DINOv2 is indeed excellently suited for visual AD.
In addition, ViT-based architectures like DINOv2 easily allow handling images with varying resolutions. In this context, we find that operating at a resolution of 448 gives the best trade-off between performance and inference time, but even higher detection AUROCs are possible at higher resolutions.
We also evaluated our pipeline with DINOv2 at different distillation sizes (ViT-S, ViT-B, ViT-L). The full results are given in \Cref{App:Ablation-Size}, \Cref{fig:ablation_model_size}. 
We observe no considerable differences, in particular, also larger backbones give state-of-the-art results on MVTec-AD (all $k$) and VisA ($k \ge 8$).
Interestingly, larger architectures do not necessarily translate to higher performance.
The smallest model demonstrates the best performance on MVTec-AD, which primarily consists of single objects and less complex scenes. 
In contrast, the larger architecture sizes perform better on VisA, which often involves multiple and more complex objects. This suggests the importance of achieving the `just right' level of abstraction for the specific task, particularly in the few-shot regime.

\section{Conclusion}
\label{sec:Discussion}
This paper proposes a vision-only approach for one- and few-shot anomaly detection.
Our method is based on similarities between patch representations extracted by DINOv2 \cite{oquab2024dino2}.
We carefully design means to populate the nominal memory bank with diverse and relevant features while minimizing false positives by utilizing zero-shot segmentation and simple data augmentation.
Industrial settings require fast deployment, easy debugging and error correction, and rapid adaptation for covariance shifts in the normal data distribution.
Our pipeline caters to these requirements through its simplicity and computational efficiency.
The proposed method, \method{}, achieves state-of-the-art results on MVTec-AD and rivals all competing methods on VisA while outperforming all other training-free approaches. 
Thus, our approach achieves the best of both worlds: improved performance \textit{and} reduced inference time, especially compared to more complex vision-language methods.
Its simplicity and strong performance render \method{} an excellent candidate for practitioners for industrial anomaly detection and an effective baseline for assessing few-shot and even full-shot anomaly detection in ongoing research.

\paragraph{Follow-up research directions}
The specific pipeline exemplified in the paper focuses on simplicity and high throughput. However, the individual parts of our pipeline can easily be exchanged for more sophisticated alternatives.
For example, our simple masking approach could be replaced by more specialized and adaptive masking techniques (which may also be relevant to other methods building on DINOv2), and
the simple upsampling and smoothing approach could be substituted by more sophisticated methods like \cite{fu2024featup}.
It would be interesting to see if that leads to further improvements in anomaly detection and localization, thereby further reducing the gap between few- and full-shot anomaly detectors. We also plan to improve the batched zero-shot performance of \method{}.

\paragraph{Acknowledgement} 
The authors acknowledge funding from TRR 391 \textit{Spatio-temporal Statistics for the Transition of Energy and Transport} by the German Research Foundation (DFG).

{\small
\bibliographystyle{plain}
\bibliography{bibliography}
}


\clearpage

\appendix

\twocolumn[ 
    \begin{center}
        {\Large \textbf{Supplementary Material}} 

        \vspace{2mm}
    \end{center}
]

\section{Detailed experimental results}
\label{App:DetailedResults}

\begin{figure}[!htb]
    \centering
    \begin{tabular}{*{8}{>{\centering\arraybackslash}p{0.095\textwidth}}}
        \footnotesize Query & \footnotesize GT & \footnotesize 1-shot & \footnotesize 8-shot &
    \end{tabular}
    \includegraphics[width=0.495\textwidth,valign=t]{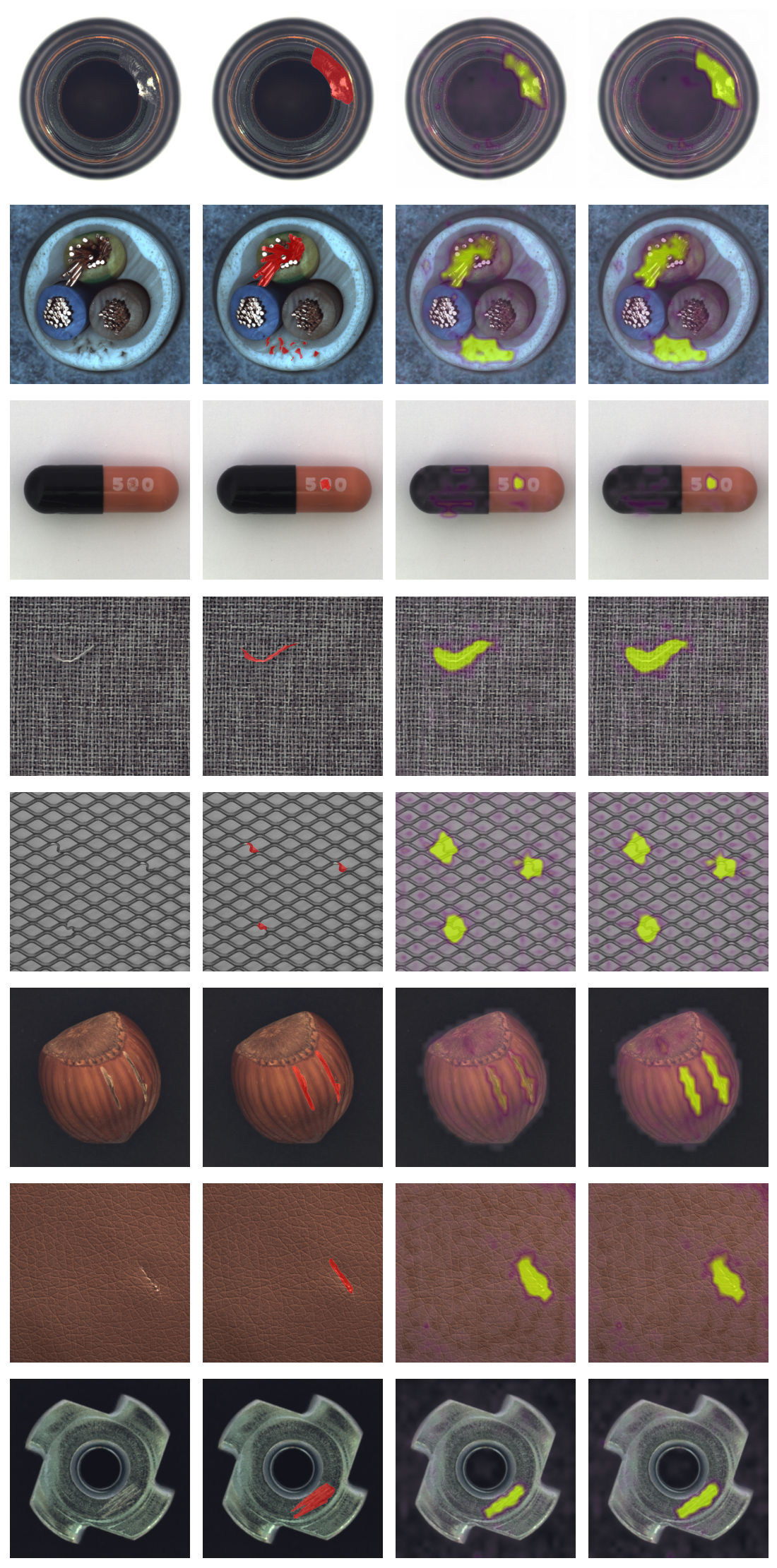}
    \caption{\textbf{Examples -- MVTec-AD (1/2)}. Depicted are, from left to right, a test sample per category (Query), the ground truth anomaly annotation (GT), and the predicted anomaly map from \method{}-S (448) in the 1- and 8-shot settings. 
    The color coding is normalized by the max. score over `good' test samples.}
    \label{fig:MVTec_examples}
\end{figure}

\begin{figure}[!bthp]
\vspace{-2.2mm}
    \centering
    \begin{tabular}{*{8}{>{\centering\arraybackslash}p{0.095\textwidth}}}
        \footnotesize Query & \footnotesize GT & \footnotesize 1-shot & \footnotesize 8-shot &
    \end{tabular}
    \includegraphics[width=0.495\textwidth,valign=t]{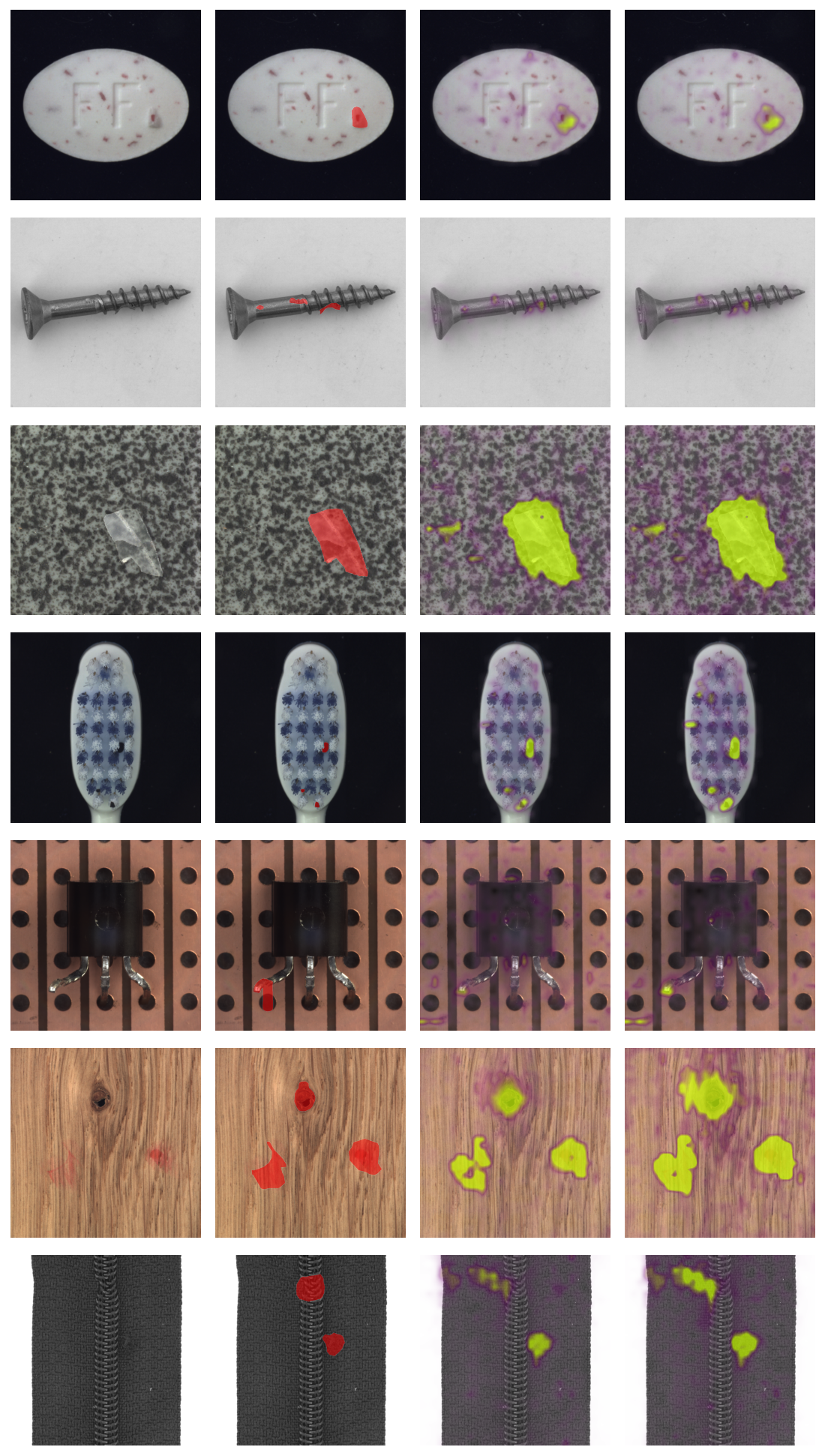}
    \caption{\textbf{Examples -- MVTec-AD (2/2)}. See \cref{fig:MVTec_examples} for a description.}
    \label{fig:MVTec_examples2}
    \vspace{-2mm}
\end{figure}

\begin{figure}[!htb]
    \centering
    \begin{tabular}{*{8}{>{\centering\arraybackslash}p{0.095\textwidth}}}
        \footnotesize Query & \footnotesize GT & \footnotesize 1-shot & \footnotesize 8-shot &
    \end{tabular}
    \includegraphics[width=0.495\textwidth,valign=t]{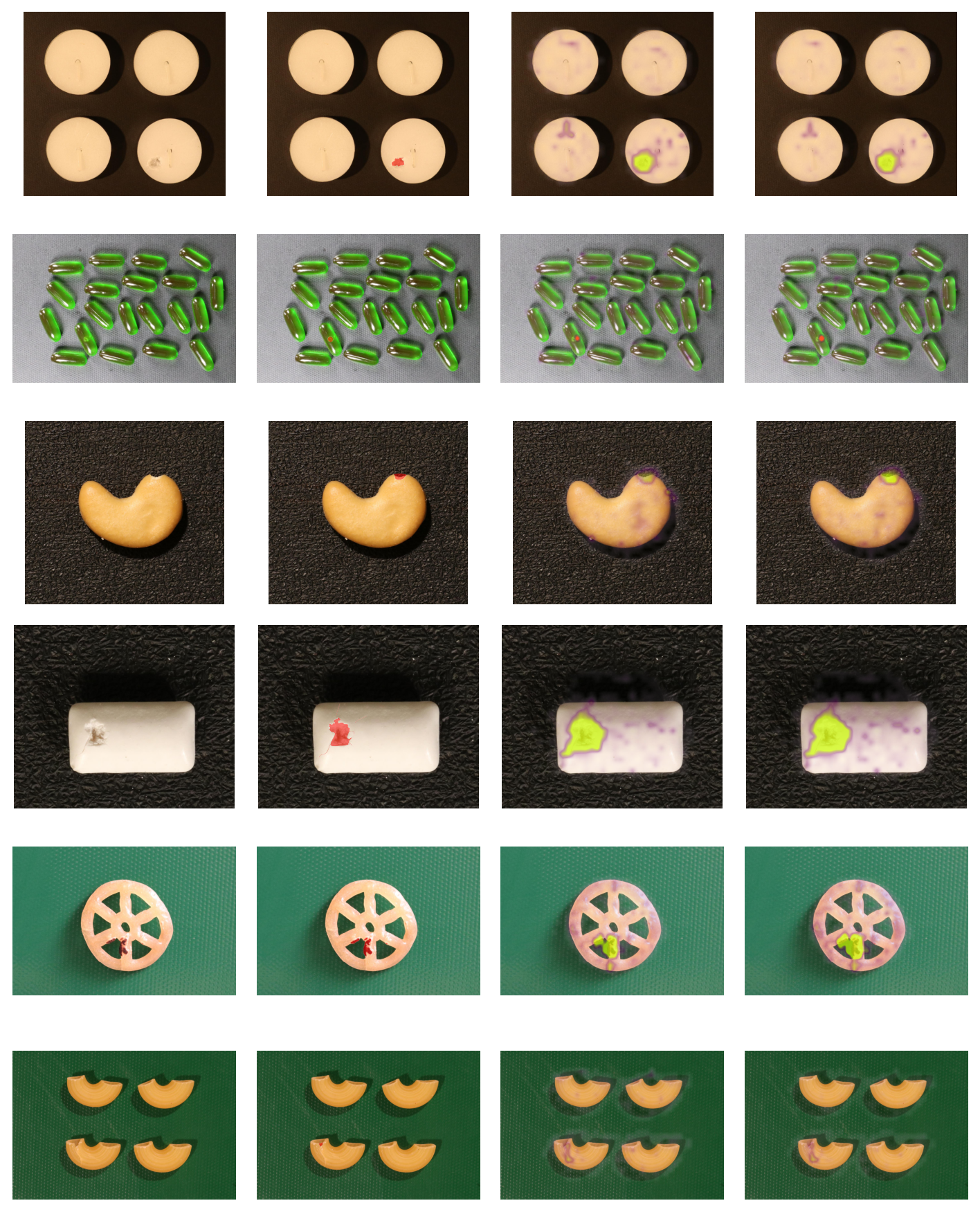}

    \caption{\textbf{Examples -- VisA (1/2)}. Depicted are, from left to right, a test sample per category (Query), the ground truth anomaly annotation (GT), and the predicted anomaly map from \method{}-S (448) in the 1- and 8-shot settings. 
    The color bar is normalized by the maximum score on the `good' test samples (per category). For the category `Capsules' (left column, second from top) we changed the color map for better visibility.
    Best viewed at a higher zoom level as some anomalies are quite small.}
    \label{fig:VisA_examples}
\end{figure}

\begin{figure}[!htb]
    \centering
    \begin{tabular}{*{8}{>{\centering\arraybackslash}p{0.095\textwidth}}}
        \footnotesize Query & \footnotesize GT & \footnotesize 1-shot & \footnotesize 8-shot &
    \end{tabular}
    \includegraphics[width=0.495\textwidth,valign=t]{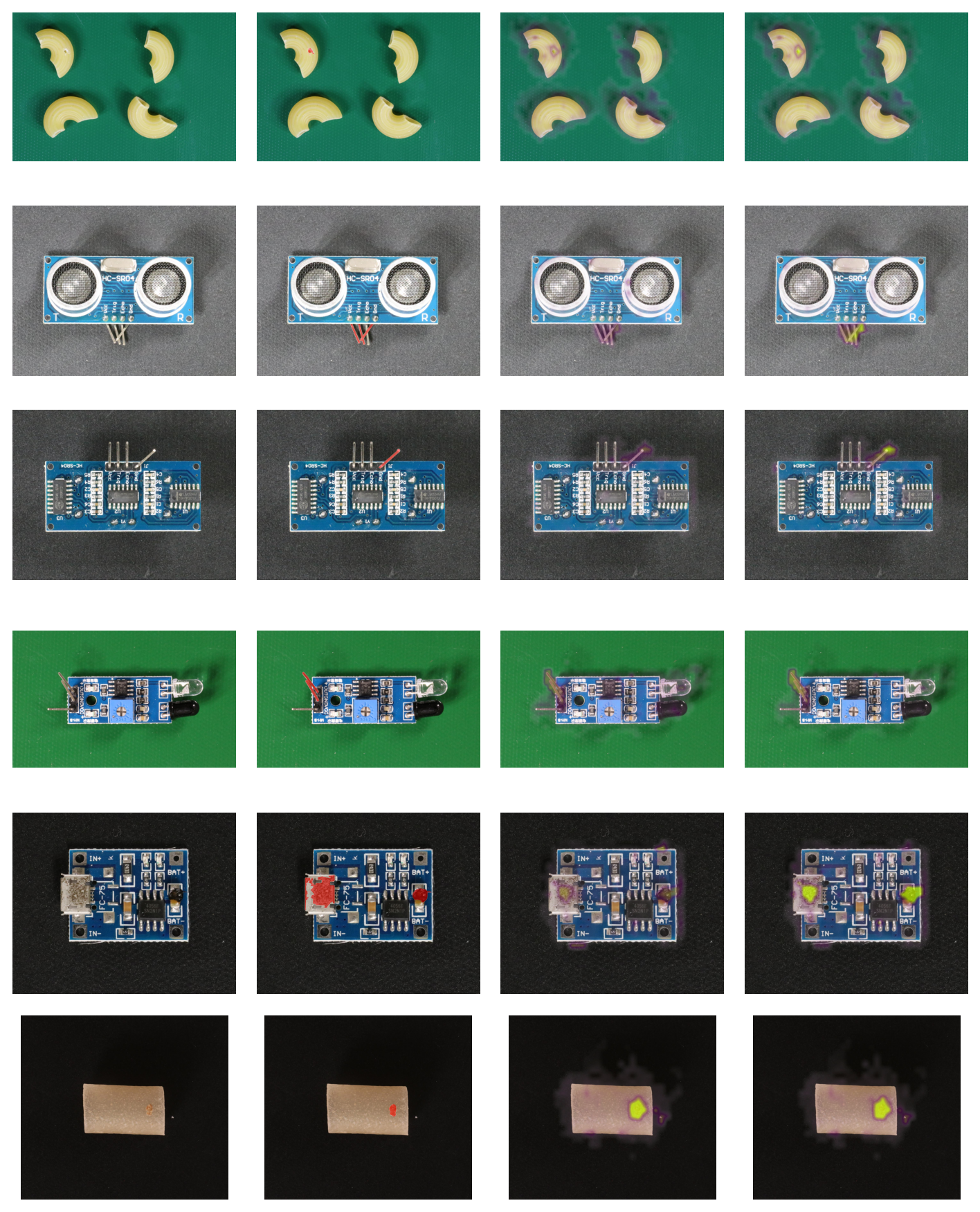}
    \caption{\textbf{Examples -- VisA (2/2)}. See \cref{fig:VisA_examples} for the description.}
    \label{fig:VisA_examples2}
\end{figure}

\noindent
Further anomaly maps, predicted by \method{}, are presented in \Cref{fig:MVTec_examples,fig:MVTec_examples2,fig:VisA_examples,fig:VisA_examples2}.
The full results per category for MVTec-AD and VisA are given in \Cref{tab:MVTec-detailed-agnostic,tab:VisA-detailed-agnostic}, respectively.

The results presented in \Cref{tab:MVTec-detailed-agnostic,tab:VisA-detailed-agnostic} show that a) anomalies in some categories are more difficult to than others, 
b), that the performance of \method{} increases across the board with more available reference samples.
In particular, more complex objects like `PCB3' and `PCB4' in VisA or `Transistor' in MVTec-AD seem to benefit the most from more available reference samples.
In addition, we see that also the variance in the reported metrics decreases with the number of nominal samples.

In this context, we observe a peculiarity of the few-shot regime (which does not occur for sufficiently populated $\mathcal{M}$).
The results crucially depend on the chosen reference image(s).\footnote{Note that this holds for all one- and few-shot methods, not only for \method{}.}
The high variances, predominantly in the 1- and 2-shot setting, for category `Capsule' for MVTec-AD, or `Cashew' or `PCB4' for VisA demonstrate this.
We discuss the potential failure cases of choosing a sub-optimal reference sample in the following section (\Cref{App:Limitations-Reference}).

Finally, we also report the full-shot performance, see \Cref{tab:full-shot}. As the diversity within $\mathcal{M}$ is sufficiently high given the large number of reference samples, we only apply masing here (no augmentations).
The results demonstrate that the performance further improves for all considered metrics.
Notably, \method{}-S (672) achieves new state-of-the-art segmentation performance measured in (AU)PRO in the full-shot setting (see \href{https://paperswithcode.com/sota/anomaly-detection-on-visa}{here}, accessed 11/26/2024).

\begin{table}[!hbt]
\caption{\textbf{Full-shot results} on MVTec-AD and VisA with AnomalyDINO-S in the default setting (no std reported as results are deterministic when all samples are considered).
}
\vspace{-2mm}
    \centering
\resizebox{\linewidth}{!}{
\begin{tabular}{@{}ccccccccc}
\toprule
\multirow{2}{*}{Dataset} & \multirow{2}{*}{Resolution} & \multicolumn{3}{c}{Detection}                                               &  & \multicolumn{3}{c}{Segmentation}                                                 \\ \cmidrule{3-9}
              &  & \multicolumn{1}{c}{AUROC} & \multicolumn{1}{c}{F1-max} & \multicolumn{1}{c}{AP}  &  & \multicolumn{1}{c}{AUROC} & \multicolumn{1}{c}{F1-max} & \multicolumn{1}{c}{PRO} \\ \midrule
\multirow{2}{*}{MVTec-AD}    &  448  & 99.3 & 98.8 & 99.7 &  &  97.9 & 61.8 &  93.9 \\
                             &  672  & 99.5 & 99.0 & 99.8 &  &  98.2 & 64.3 &  95.0 \\ 
\midrule
\multirow{2}{*}{VisA}        & 448     & 97.2 & 93.7 & 97.6 &  &  98.7 & 50.5 &  95.0  \\  
                             & 672     & 97.6 & 94.5 & 98.0 &  &  98.8 & 53.8 &  96.1 \\  
\bottomrule
\end{tabular}}
\vspace{-2mm}
\label{tab:full-shot}
\end{table}

\begin{table*}[!hbt]
\centering
\caption{\textbf{Detailed results on MVTec-AD}. 
Reported are results on anomaly detection (image-level AUROC) and segmentation (PRO) of \method{}-S (672) with default preprocessing (mean and standard deviation over three independent runs, all results in \%).}
\resizebox{1\textwidth}{!}{
    \begin{tabular}{@{}rrrrrrrrrrr@{}}
\toprule
\multirow{2}{*}{Shots}  & \multicolumn{2}{c}{1-shot}     & \multicolumn{2}{c}{2-shot}     & \multicolumn{2}{c}{4-shot}     & \multicolumn{2}{c}{8-shot}     & \multicolumn{2}{c}{16-shot}    \\ \cmidrule(l){2-11} 
           & AUROC          & PRO           & AUROC          & PRO           & AUROC          & PRO           & AUROC          & PRO           & AUROC          & PRO           \\ \midrule 

Bottle & 99.7\std{0.2} & 95.9\std{0.3} & 99.9\std{0.1} & 96.3\std{0.0} & 99.9\std{0.1} & 96.7\std{0.1} & 100.0\std{0.0} & 96.5\std{0.4} & 99.9\std{0.1} & 96.4\std{0.4} \\
Cable & 92.7\std{0.8} & 89.4\std{0.4} & 92.4\std{1.1} & 89.5\std{0.4} & 93.8\std{0.9} & 90.4\std{0.3} & 95.2\std{0.3} & 90.5\std{0.2} & 95.1\std{0.6} & 90.5\std{0.6} \\
Capsule & 90.2\std{5.5} & 97.1\std{0.3} & 89.2\std{7.9} & 97.3\std{0.6} & 95.8\std{0.5} & 97.9\std{0.1} & 95.6\std{0.5} & 97.9\std{0.1} & 95.5\std{0.6} & 98.1\std{0.1} \\
Carpet & 100.0\std{0.0} & 97.8\std{0.0} & 100.0\std{0.0} & 97.9\std{0.0} & 100.0\std{0.0} & 97.8\std{0.0} & 100.0\std{0.0} & 97.8\std{0.0} & 100.0\std{0.0} & 97.8\std{0.0} \\
Grid & 99.1\std{0.2} & 97.2\std{0.1} & 99.2\std{0.4} & 97.2\std{0.1} & 99.5\std{0.3} & 97.2\std{0.1} & 99.5\std{0.1} & 97.2\std{0.0} & 99.7\std{0.3} & 97.0\std{0.2} \\
Hazelnut & 97.5\std{2.6} & 97.4\std{0.4} & 99.6\std{0.5} & 98.0\std{0.3} & 99.8\std{0.1} & 98.0\std{0.1} & 100.0\std{0.0} & 98.1\std{0.1} & 100.0\std{0.0} & 98.1\std{0.1} \\
Leather & 100.0\std{0.0} & 97.9\std{0.1} & 100.0\std{0.0} & 97.8\std{0.0} & 100.0\std{0.0} & 97.6\std{0.1} & 100.0\std{0.0} & 97.6\std{0.1} & 100.0\std{0.0} & 97.3\std{0.2} \\
Metal nut & 99.9\std{0.1} & 94.2\std{0.0} & 100.0\std{0.0} & 94.6\std{0.2} & 100.0\std{0.0} & 95.3\std{0.1} & 100.0\std{0.0} & 95.4\std{0.3} & 100.0\std{0.0} & 95.7\std{0.1} \\
Pill & 93.7\std{0.9} & 97.3\std{0.1} & 95.4\std{0.7} & 97.5\std{0.1} & 96.0\std{0.2} & 97.6\std{0.1} & 97.2\std{0.2} & 97.7\std{0.1} & 97.9\std{0.1} & 97.8\std{0.1} \\
Screw & 93.2\std{0.3} & 93.4\std{0.4} & 93.5\std{0.8} & 94.3\std{0.4} & 92.7\std{2.3} & 94.5\std{0.9} & 93.5\std{1.1} & 95.2\std{0.5} & 94.7\std{0.9} & 95.9\std{0.3} \\
Tile & 100.0\std{0.0} & 88.0\std{0.2} & 100.0\std{0.0} & 87.6\std{0.4} & 100.0\std{0.0} & 87.1\std{0.4} & 100.0\std{0.0} & 86.7\std{0.2} & 100.0\std{0.0} & 86.0\std{0.5} \\
Toothbrush & 97.4\std{0.5} & 94.0\std{0.8} & 98.1\std{1.0} & 94.7\std{0.3} & 97.5\std{0.6} & 94.7\std{0.3} & 97.7\std{1.6} & 95.1\std{0.8} & 98.1\std{1.8} & 95.8\std{1.0} \\
Transistor & 90.9\std{1.2} & 67.3\std{2.1} & 89.4\std{4.6} & 68.4\std{3.1} & 93.2\std{2.2} & 70.6\std{1.4} & 96.2\std{1.3} & 75.3\std{1.9} & 97.6\std{0.3} & 78.2\std{1.4} \\
Wood & 98.0\std{0.2} & 94.7\std{0.1} & 98.0\std{0.1} & 94.6\std{0.0} & 97.9\std{0.2} & 94.6\std{0.1} & 98.3\std{0.4} & 94.4\std{0.3} & 98.3\std{0.6} & 94.2\std{0.4} \\
Zipper & 97.4\std{0.9} & 89.2\std{1.2} & 98.9\std{0.4} & 90.2\std{0.4} & 99.0\std{0.4} & 91.2\std{0.3} & 99.6\std{0.1} & 91.1\std{0.4} & 99.6\std{0.3} & 91.7\std{0.5} \\ \midrule
Mean & 96.6\std{0.4} & 92.7\std{0.1} & 96.9\std{0.7} & 93.1\std{0.2} & 97.7\std{0.2} & 93.4\std{0.1} & 98.2\std{0.2} & 93.8\std{0.1} & 98.4\std{0.1} & 94.0\std{0.1} \\ \bottomrule
\end{tabular}

}
\label{tab:MVTec-detailed-agnostic}
\end{table*}

\begin{table*}[!ht]
\caption{\textbf{Detailed results on VisA}. 
Reported are results on anomaly detection (image-level AUROC) and segmentation (PRO) of \method{}-S (672) with default preprocessing (mean and standard deviation over three independent runs, all results in \%).}
\centering
 \resizebox{1\textwidth}{!}{
\begin{tabular}{@{}rcccccccccc@{}}
\toprule
\multirow{2}{*}{Shots} & \multicolumn{2}{c}{1-shot}                          & \multicolumn{2}{c}{2-shot}                          & \multicolumn{2}{c}{4-shot}                          & \multicolumn{2}{c}{8-shot}                          & \multicolumn{2}{c}{16-shot}                       \\ \cmidrule(l){2-11} 
                       & \multicolumn{1}{c}{AUROC} & \multicolumn{1}{c}{PRO} & \multicolumn{1}{c}{AUROC} & \multicolumn{1}{c}{PRO} & \multicolumn{1}{c}{AUROC} & \multicolumn{1}{c}{PRO} & \multicolumn{1}{c}{AUROC} & \multicolumn{1}{c}{PRO} & \multicolumn{1}{c}{AUROC} & \multicolumn{1}{c}{PRO}\\ \midrule

Candle & 87.9\std{0.3} & 96.8\std{0.4} & 89.4\std{3.0} & 97.0\std{0.2} & 91.3\std{2.9} & 97.2\std{0.1} & 93.5\std{1.2} & 97.3\std{0.2} & 94.5\std{0.5} & 97.6\std{0.2} \\
Capsules & 98.4\std{0.5} & 95.1\std{0.7} & 98.9\std{0.1} & 95.5\std{0.2} & 99.2\std{0.1} & 96.3\std{0.4} & 99.2\std{0.1} & 96.7\std{0.2} & 99.2\std{0.2} & 97.2\std{0.3} \\
Cashew & 86.1\std{3.6} & 96.1\std{0.9} & 89.4\std{3.8} & 96.7\std{0.7} & 94.5\std{0.7} & 97.4\std{0.5} & 95.3\std{0.6} & 97.3\std{0.2} & 96.0\std{0.3} & 97.3\std{0.1} \\
Chewinggum & 98.0\std{0.4} & 92.0\std{1.0} & 98.6\std{0.4} & 92.9\std{0.3} & 98.8\std{0.2} & 93.0\std{0.1} & 98.8\std{0.2} & 93.1\std{0.3} & 98.8\std{0.2} & 93.0\std{0.3} \\
Fryum & 94.8\std{0.5} & 93.2\std{0.2} & 96.5\std{0.2} & 93.9\std{0.3} & 97.0\std{0.1} & 94.5\std{0.4} & 97.6\std{0.4} & 94.9\std{0.3} & 97.9\std{0.2} & 95.1\std{0.0} \\
Macaroni1 & 87.5\std{1.1} & 97.5\std{0.3} & 87.5\std{0.9} & 97.9\std{0.3} & 89.5\std{1.4} & 98.3\std{0.2} & 90.1\std{1.7} & 98.6\std{0.2} & 90.4\std{1.1} & 98.7\std{0.1} \\
Macaroni2 & 62.2\std{4.3} & 92.0\std{0.7} & 66.9\std{1.9} & 93.0\std{0.4} & 70.0\std{1.7} & 93.9\std{0.8} & 74.9\std{0.4} & 95.0\std{0.6} & 77.6\std{0.8} & 95.7\std{0.3} \\
PCB1 & 91.5\std{2.0} & 92.6\std{0.2} & 91.2\std{2.7} & 92.5\std{0.5} & 94.0\std{2.1} & 93.3\std{0.5} & 95.5\std{0.5} & 93.9\std{0.2} & 96.8\std{0.7} & 94.2\std{0.2} \\
PCB2 & 84.8\std{1.2} & 89.9\std{0.2} & 88.1\std{2.5} & 90.7\std{0.3} & 91.1\std{1.7} & 91.4\std{0.2} & 92.6\std{0.3} & 92.0\std{0.1} & 93.2\std{0.1} & 92.5\std{0.2} \\
PCB3 & 84.9\std{3.3} & 88.5\std{1.3} & 89.4\std{3.8} & 90.8\std{0.5} & 94.3\std{0.4} & 91.7\std{0.4} & 95.6\std{0.2} & 93.1\std{0.3} & 96.5\std{0.3} & 93.9\std{0.3} \\
PCB4 & 79.9\std{13.7} & 78.5\std{6.8} & 87.4\std{11.3} & 82.0\std{6.1} & 96.2\std{2.6} & 84.1\std{1.5} & 98.0\std{0.3} & 87.9\std{2.3} & 99.0\std{0.4} & 90.4\std{1.8} \\
Pipe fryum & 92.7\std{2.7} & 98.0\std{0.0} & 93.3\std{1.2} & 97.9\std{0.2} & 94.6\std{1.9} & 97.8\std{0.1} & 95.0\std{1.6} & 97.6\std{0.1} & 97.2\std{0.9} & 97.7\std{0.1} \\ \midrule
Mean & 87.4\std{1.2} & 92.5\std{0.5} & 89.7\std{1.3} & 93.4\std{0.6} & 92.6\std{0.9} & 94.1\std{0.1} & 93.8\std{0.3} & 94.8\std{0.2} & 94.8\std{0.2} & 95.3\std{0.2} \\ \bottomrule
\end{tabular}
}
\label{tab:VisA-detailed-agnostic}
\vspace{-4mm}
\end{table*}

\section{Limitations and failure cases}
\label{App:Limitations}
The proposed method relies on similarities to patch representations captured in $\mathcal{M}$.
Therefore, we can only expect the model to detect those anomalies caused by regions in the test images that are particularly different from patches of the given reference sample(s).
Our experiments reveal that this may lead to some specific failure cases.

\subsection{Semantic Anomalies}
\begin{figure}[!th]
    \centering
    \begin{subfigure}[t]{0.3\linewidth}
        \centering
        \includegraphics[width=\linewidth]{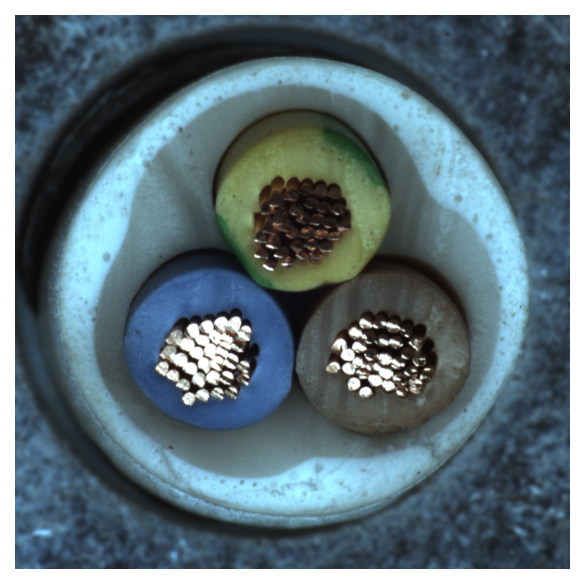}
        \caption{Reference sample (No anomalies)}
    \end{subfigure}
    \hfill
    \begin{subfigure}[t]{0.3\linewidth}
        \centering
        \includegraphics[width=\linewidth]{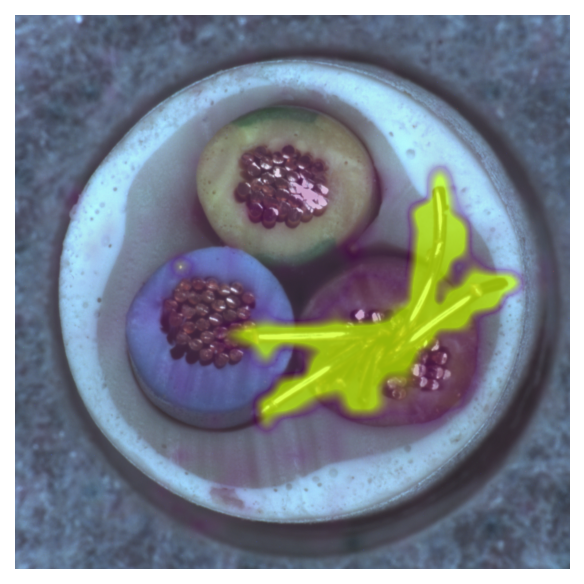}
        \caption{Anomaly type `Bent Wire'}
        \label{fig:sensory_anomaly}
    \end{subfigure}
    \hfill
    \begin{subfigure}[t]{0.3\linewidth}
        \centering
        \includegraphics[width=\linewidth]{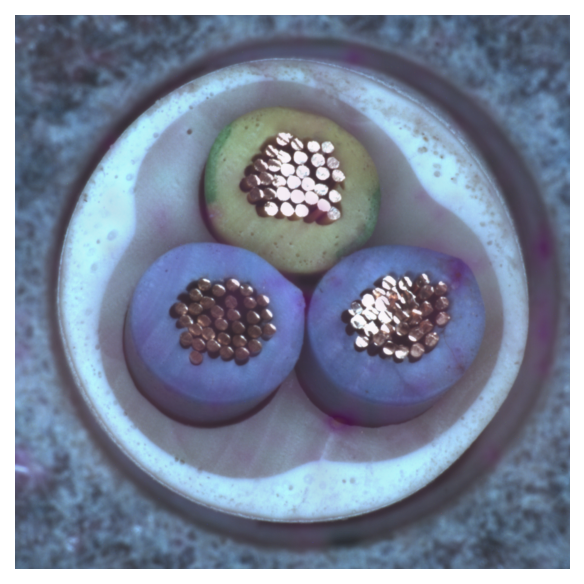}
        \caption{Anomaly type `Cable Swap'}
        \label{fig:semantic_anomaly}
    \end{subfigure}
    \caption{\textbf{Example of a semantic anomaly} in category `Cable' (MVTec-AD). 
    Depicted are a nominal reference sample, a sensory anomaly (Figure~\ref{fig:sensory_anomaly}), and a semantic anomaly (Figure~\ref{fig:semantic_anomaly}) with anomaly maps predicted by \method{} (1-shot).}
    \label{fig:cable_swap}
\end{figure}

The first relates to the distinction between low-level sensory anomalies and high-level semantic anomalies.
Semantic anomalies might be present because a logical rule or a specific semantic constraint is violated. In contrast, \method{} is developed with low-level sensory anomalies in mind.
Consider, for instance, the anomalies shown in Figure~\ref{fig:cable_swap}. 

While the cable with anomaly type `Bent Wire' (Figure~\ref{fig:sensory_anomaly}) contains patches that cannot be well matched to patches of the reference image, this does not apply to the anomaly `Cable Swap' (Figure~\ref{fig:semantic_anomaly}). 
The latter shows a \textit{semantic anomaly} with two blue wires, while a nominal image of a `Cable' should depict all three different cable types.
As a result, all test patches of \Cref{fig:sensory_anomaly} can be matched well with reference patches in $\mathcal{M}$, and thus, \method{} does not detect this anomaly type.  
Evaluating the detection performance for this specific failure case, `Cable Swap', our proposed method essentially performs on chance level, giving a detection AUROC of 50.2\% ($\pm$ 4.9\%).

\subsection{The importance of informative reference samples}
\label{App:Limitations-Reference}

\begin{figure}[!bht]   
\vspace{-2mm}
    \begin{subfigure}[t]{0.45\textwidth}
        \centering
        \includegraphics[width=0.42\linewidth]{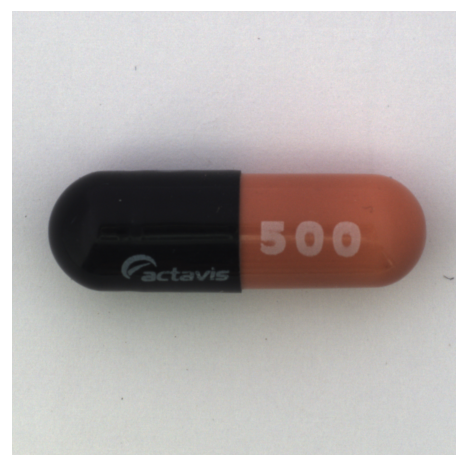}
        \includegraphics[width=0.42\linewidth]{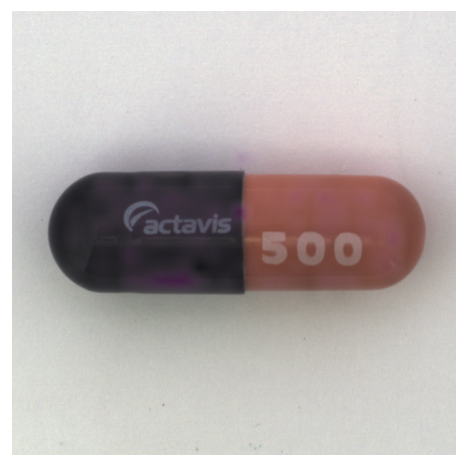}
        \caption{Reference sample (left) and estimated anomaly map of `good' test sample (right).}
    \end{subfigure}
    \hfill
    \begin{subfigure}[t]{0.45\textwidth}
        \centering
        \includegraphics[width=0.42\linewidth]{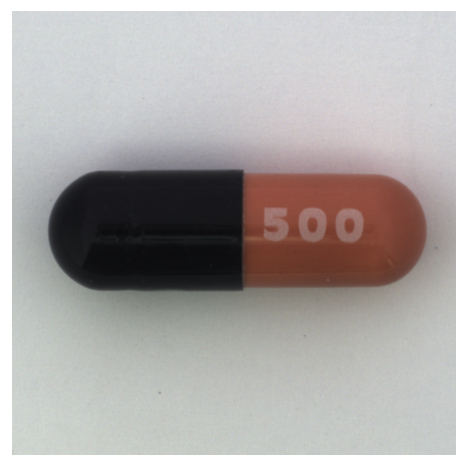}
        \includegraphics[width=0.42\linewidth]{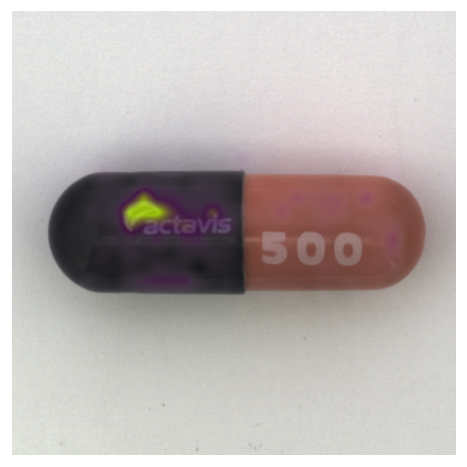}
        \caption{Reference sample (left) and estimated anomaly map of `good' test sample (right).}
        \label{fig:sample_variation_no_text}
    \end{subfigure}
    \caption{\textbf{Example of an uninformative reference sample} in category `Capsule' (MVTec-AD). 
    Some reference samples do not resemble the full concept of normality (here, the sample in Figure~\ref{fig:sample_variation_no_text} does not show the text on the capsule, i.e., any nominal sample with text be considered anomalous).
    Anomaly maps predicted by \method{} based on the depicted reference sample.}
    \label{fig:failure_case_capsule}
\end{figure}

The second failure case occurs if the reference sample(s) does not resemble all concepts of normality, therefore $\mathcal{M}$ does not capture all variations of the nominal distribution $p_\mathrm{norm}$.
As an illustration, consider the nominal samples of `Capsule' in MVTec-AD, which may be rotated in such a way that the imprinted text is hidden (see \Cref{fig:sample_variation_no_text}).
Therefore, parts of the text of a nominal test sample may be falsely recognized as anomalies.
Due to the strong dependency on a suitable reference sample, we observe higher AD variances for some products in the one-shot setting, as shown in \Cref{tab:MVTec-detailed-agnostic,tab:VisA-detailed-agnostic}.
We like to remark, that this is only relevant to the few-shot setting, and with an increasing number of reference samples (and higher diversity of nominal patches in $\mathcal{M}$), the variance decreases notably.

\begin{figure*}[!ht]
\centering
    \begin{subfigure}[t]{0.9\textwidth}
        \centering
        \includegraphics[width=\textwidth]{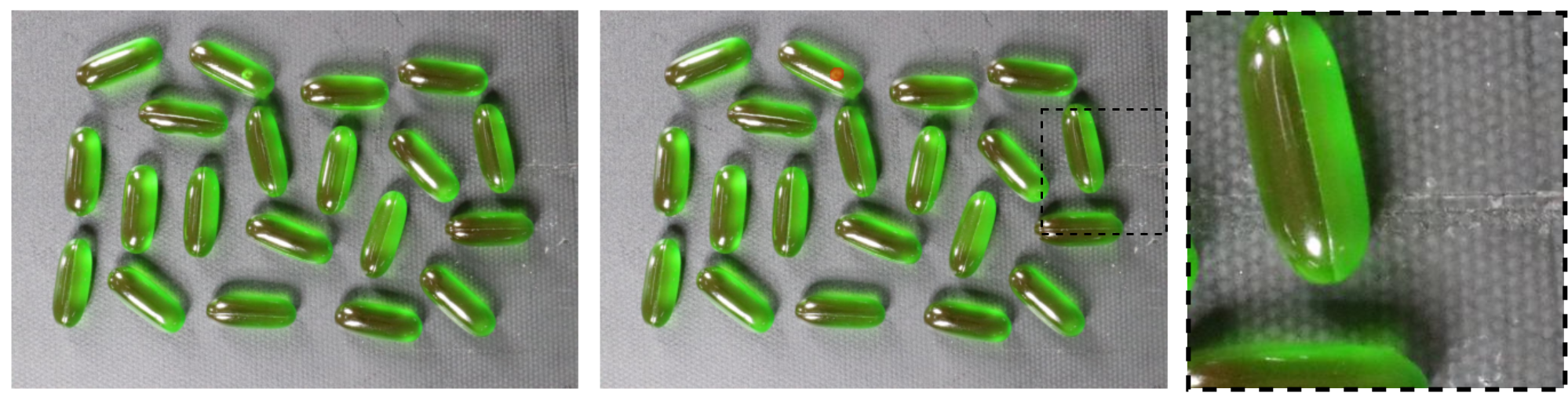}
        \caption{Test sample (left), the same test sample with ground-truth anomaly annotation (center), and a magnified view of a region with strong background artifacts (right).}
    \end{subfigure}
    \hfill
    \begin{subfigure}[t]{0.9\textwidth}
        \centering
        \includegraphics[width=0.49\linewidth]{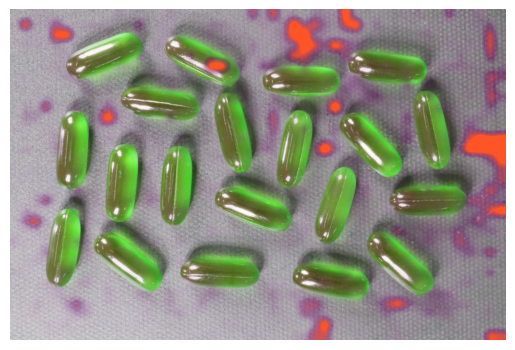}
        \hfill
        \includegraphics[width=0.49\linewidth]{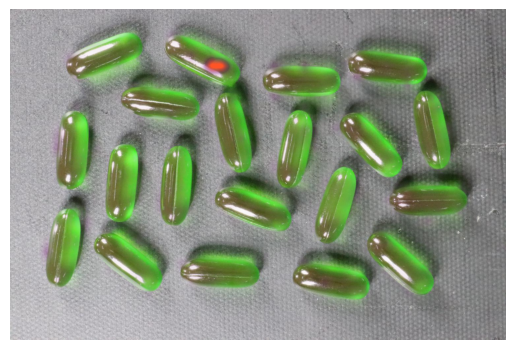}
        \caption{Anomaly map predicted by \method{} (1-shot) without masking (left) and with masking (right).}
    \end{subfigure}

    \caption{\textbf{Visualization of the effect of masking} in the presence of high background noise for the category `Capsules' in VisA (1-shot).
    As in \Cref{fig:VisA_examples} we depict the anomaly map for `Capsules' using a different colormap (red instead of yellow) for better visibility. 
    Best viewed on a higher zoom level.}
    \label{fig:masking-background-noise}
\end{figure*}

\section{Ablation Study}
\label{App:Ablation}

\subsection{Preprocessing}
\label{App:Ablation-Preprocessing}

As discussed in \Cref{sec:Method}, we consider two potential preprocessing steps in our pipeline: masking and rotations.
We mask out irrelevant background patches, whenever the zero-shot segmentation of DINOv2 captures the object correctly (see \Cref{fig:Preprocessing_Masking_Test}). 
Discarding background patches helps to mitigate the problem of potential background noise, thereby reducing the number of false positives.
A representative example, here from the category `Capsules' from the VisA dataset, is depicted in \Cref{fig:masking-background-noise}.
Without masking, the method would correctly predict the depicted sample to show an anomaly, but for the wrong reason (high anomaly scores caused by background noise/contamination). In contrast, the irrelevant background areas are discarded by our masking approach, such that the anomalous region is correctly identified.
We conclude that masking is essential to faithfully detect anomalies whenever higher variations in the background are expected---and as the issue of low variations is particularly pressing in the few-shot regime, caused by the minimal amount of nominal reference samples, suitable masking can significantly boost the performance in these cases. This is showcased by the superior anomaly localization performance 
A quantitative analysis is provided in the next paragraphs.

In addition to masking (which decreases the size of $\mathcal{M}$), we consider augmenting the reference sample with rotations (which increases the size and diversity of $\mathcal{M}$). Here we distinguish the `agnostic' scenario, where we do not know a priori about potential rotations, and augment by default. 
Whenever we know about potential rotations of reference or test samples, we can deactivate the augmentation to reduce the size of the memory bank $\mathcal{M}$, the time to construct $\mathcal{M}$, as well as the test time itself.

The preprocessing decision per category for MVTec-AD and VisA, inferred (solely) from the first nominal reference sample in $X_\mathrm{ref}$ (to comply with the one-shot setting), are given in \Cref{tab:Preprocessing_Default}.

\begin{table}[!b]
\centering
\caption{\textbf{Default preprocessing steps for MVTec-AD and VisA}.
We do not mask textures, as indicated by (T).
In addition, we do not apply masking when the masking test on the first train sample failed, as indicated by (MT).  See \Cref{sec:Method} and \Cref{fig:Preprocessing_Masking_Test} for further discussion and visualization.}
\resizebox{\linewidth}{!}{
\begin{tabular}[t]{@{}rccc@{}}
\toprule
\multicolumn{1}{c}{\textbf{MVTec-AD}} & Mask?                & \multicolumn{2}{c}{Rotation?} \\ \cmidrule{2-4}
Object               & \multicolumn{1}{l}{} & informed      & agnostic      \\ \midrule 
Bottle               & \xmark (MT)          & \xmark        & \cmark        \\
Cable                & \xmark (MT)          & \xmark        & \cmark        \\
Capsule              & \cmark               & \xmark        & \cmark        \\
Carpet               & \xmark (T)           & \xmark        & \cmark        \\
Grid                 & \xmark (T)           & \xmark        & \cmark        \\
Hazelnut             & \cmark               & \cmark        & \cmark        \\
Leather              & \xmark (T)           & \xmark        & \cmark        \\
Metal nut            & \xmark (MT)          & \xmark        & \cmark        \\
Pill                 & \cmark               & \xmark        & \cmark        \\
Screw                & \cmark               & \cmark        & \cmark        \\
Tile                 & \xmark (T)           & \xmark        & \cmark        \\
Toothbrush           & \cmark               & \xmark        & \cmark        \\
Transistor           & \xmark (MT)          & \xmark        & \cmark        \\
Wood                 & \xmark (T)           & \xmark        & \cmark        \\
Zipper               & \xmark (MT)          & \xmark        & \cmark        \\ \bottomrule
\end{tabular} \quad 


\begin{tabular}[t]{@{}rccc@{}}
\toprule
\multicolumn{1}{c}{\textbf{VisA}}  & Mask?                & \multicolumn{2}{c}{Rotation?} \\ \cmidrule{2-4}
Object & \multicolumn{1}{l}{} & informed      & agnostic      \\  \midrule
Candle               & \cmark               & \xmark        & \cmark        \\
Capsules             & \cmark               & \xmark        & \cmark        \\
Cashew               & \cmark               & \xmark        & \cmark        \\
Chewinggum           & \cmark               & \xmark        & \cmark        \\
Fryum                & \cmark               & \xmark        & \cmark        \\
Macaroni1            & \cmark               & \xmark        & \cmark        \\
Macaroni2            & \cmark               & \xmark        & \cmark        \\
PCB1                 & \cmark               & \xmark        & \cmark        \\
PCB2                 & \cmark               & \xmark        & \cmark        \\
PCB3                 & \cmark               & \xmark        & \cmark        \\
PCB4                 & \cmark               & \xmark        & \cmark        \\
Pipe fryum           & \cmark               & \xmark        & \cmark       \\ \bottomrule
\end{tabular}}
\label{tab:Preprocessing_Default}
\end{table}

\begin{figure*}[!htb]
    \centering
    \includegraphics[width=1.0\linewidth]{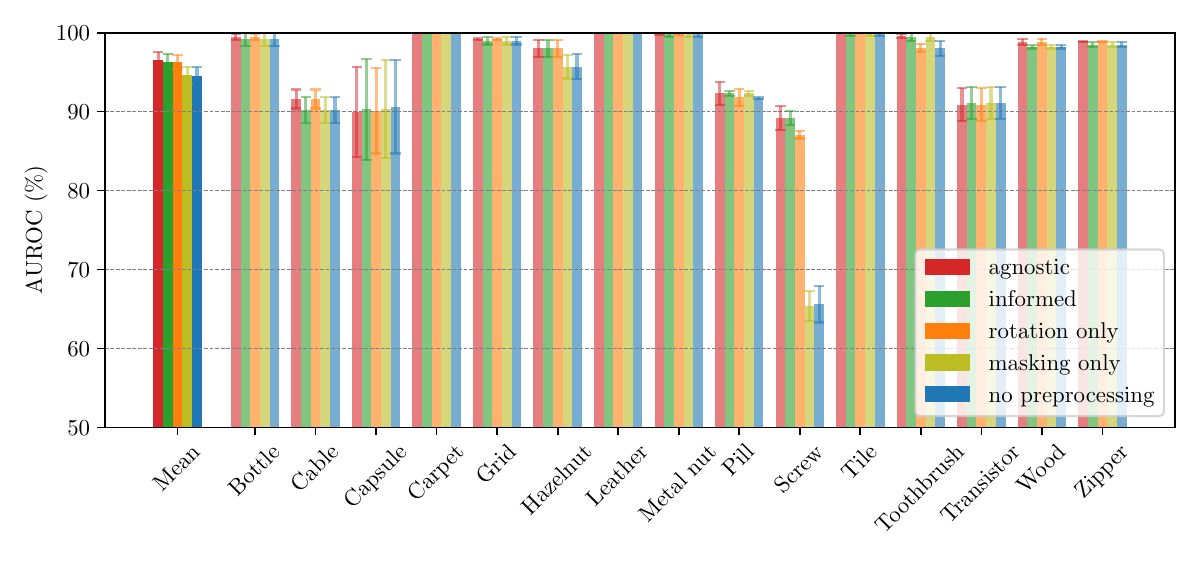}
    \vspace{-8mm}
    \caption{\textbf{Effect of preprocessing on MVTec-AD}.
    Anomaly detection of \method{}-S (448) in the 1-shot setting for different choices of the preprocessing pipeline (detection AUROC on image-level in \%, mean and standard deviation over three independent runs).}
    \label{fig:ablation_preprocessing_MVTec}
\end{figure*}

\begin{figure*}[!hbt]
    \centering
    \vspace{-2mm}
    \includegraphics[width=1.0\linewidth]{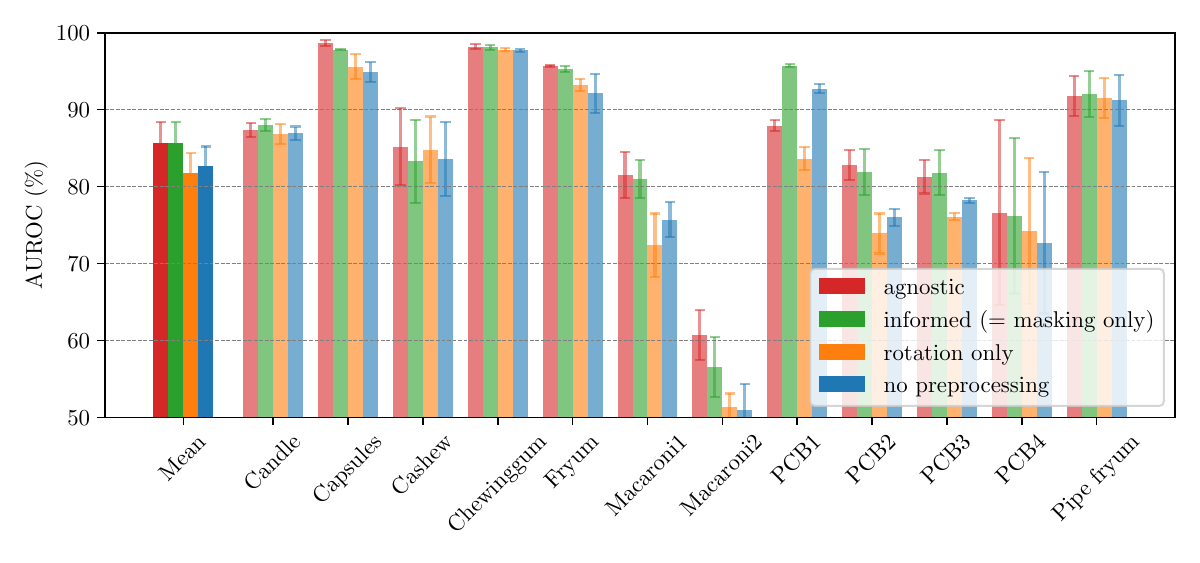}
    \vspace{-6mm}
    \caption{\textbf{Effect of preprocessing on VisA.}
    Anomaly detection with \method{}-S (448) in the 1-shot setting for different choices of the preprocessing pipeline (detection AUROC on image-level in \%, mean and standard deviation over three independent runs). For VisA, the `informed' scenario is equivalent to only applying masking (all categories), while `agnostic' is equivalent to masking and augmentations (all categories), see \Cref{tab:Preprocessing_Default}.}
    \label{fig:ablation_preprocessing_VisA}
    \vspace{-2mm}
\end{figure*}

\paragraph{Effect of preprocessing on detection performance}

As discussed in \Cref{sec:Method} (and in the previous paragraph), suitable means to fill $\mathcal{M}$ and preprocess test samples, influence the detection performance. 
The results per object for MVTec-AD and VisA are given in \Cref{fig:ablation_preprocessing_MVTec,fig:ablation_preprocessing_VisA}, respectively.

\paragraph{Rotation}
Increasing the diversity of nominal patch representations in $\mathcal{M}$ by rotating the reference sample can significantly improve the detection performance.
Consider, e.g., the category `Screw' in MVTec-AD, where the detection AUROC can be boosted from 65.6\% to 89.2\% in the 1-shot setting. This is intuitive, as the test samples of `Screw' are taken from various angles. With sufficiently many reference samples, such data augmentation is not necessary anymore, but for the few-shot setting, we see major improvements.

\begin{figure}
\centering 
    \includegraphics[width=\linewidth]{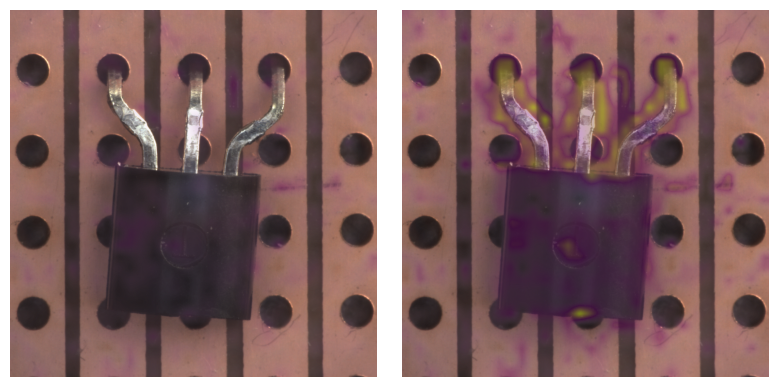} 
    \caption{\textbf{Rotations as anomalies} (`misplaced' transistor from MVTec-AD). 
    The left anomaly map is estimated from a reference sample \textit{with} rotations (`agnostic'), and the anomaly is not detected.
    In contrast, the right anomaly map is based on a reference sample \textit{without} rotations (`informed'), and the anomaly is successfully detected. This example highlights the importance of a carefully designed preprocessing pipeline for each object.}
    \label{fig:Transistor_Rotation}
\end{figure}
The same holds---although to a lesser extent---for the categories `Hazelnut', 'Cable', and `Wood' (all MVTec-AD).
However, we also observe that rotations of the reference sample can also \emph{decrease} the detection performance in some categories, namely `PCB1/2/3' and `Macaroni1' (all VisA), and to a small extent also `Transistor' (MVTec-AD).

\begin{figure}[!ht]
    \centering
    \includegraphics[width=\linewidth]{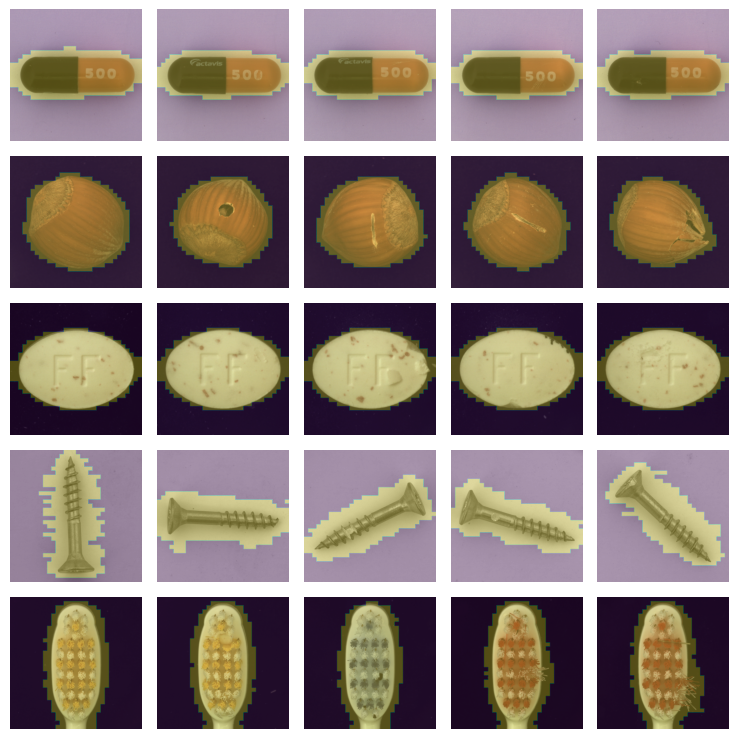}
    \caption{\textbf{Masking examples from MVTec-AD} for all categories that passed the masking test (on a single train sample, see \Cref{tab:Preprocessing_Default}).}
    \label{fig:MaskingExamplesMVTec}
    \vspace{-2mm}
\end{figure}

We attribute this to the fact the specific anomalies for these printed circuit boards contain rotated or bent connectors (see the PCB examples in \Cref{fig:VisA_examples}, right column).
And rotations of nominal samples may falsely reduce the distance between those patches depicting bent connectors and the nominal memory bank $\mathcal{M}$. 

Such a failure case based on a suboptimal preprocessing decision is depicted in \Cref{fig:Transistor_Rotation}.
The anomaly refers to a rotated transistor (anomaly type `misplaced'), and when (falsely) rotating the reference sample, such anomalies will not be detected---in contrast to the `informed' preprocessing (right side of \Cref{fig:Transistor_Rotation}). This highlights the importance of carefully designing the pre- and postprocessing pipeline for each object/product considered.

\begin{figure}[!ht]
    \centering
    \includegraphics[width=\linewidth]{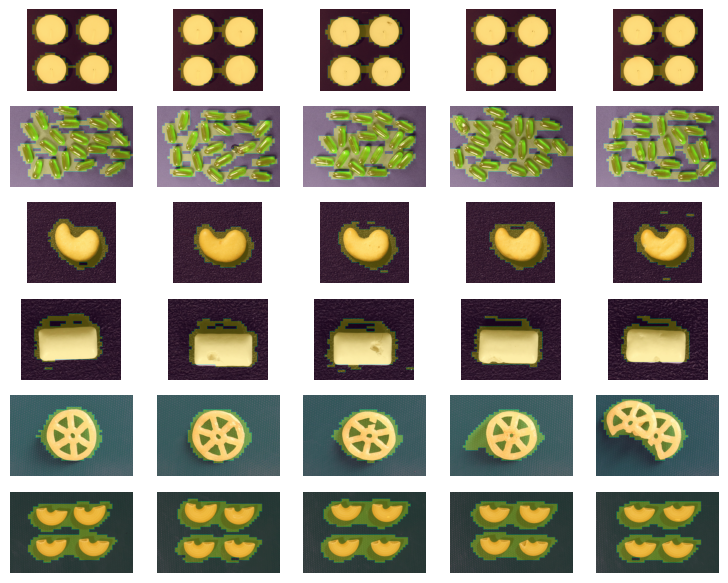}
    \includegraphics[width=\linewidth]{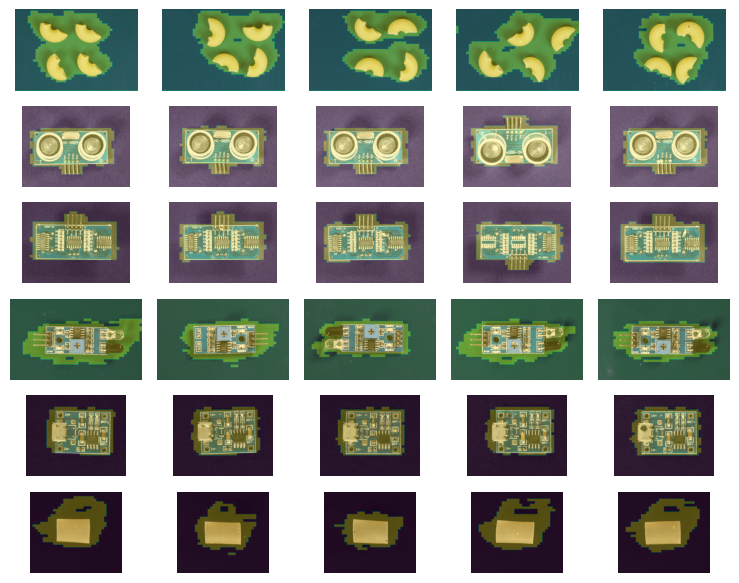}

    \caption{\textbf{Masking examples from VisA}.}
    \label{fig:MaskingExamplesVisA}
    \vspace{-2mm}
\end{figure}


\paragraph{Masking}
We utilize the zero-shot masking capabilities of DINOv2 to keep the overhead for this preprocessing step minimal. By applying a threshold to the first principal component \cite{oquab2024dino2}, we can typically distinguish between the background and the foreground. 

We also employ a straightforward rule-based strategy to enhance the robustness and generalizability of the PAC-based masking in industrial settings: ensuring that the center crop of the image is predominantly occupied by the object of interest. This minor adjustment is necessary because industrial images often differ significantly from the data on which DINOv2 was originally trained.
In addition, we use dilation and morphological closing to improve the quality of the mask. In our experiments, we applied masking only to the test samples as the size of $\mathcal{M}$ did not matter in the few-shot regime.
Across the board, we see that the proposed masking technique improves the detection performance---in many categories even significantly, e.g., for `Capsules', `Macaroni1/2', or `PCB1/2/3/4' (all VisA).
We leave further improvements and the exploration of more advanced masking techniques for future work.


\begin{table*}[ht]
    \centering
    \caption{\textbf{Runtime analysis} on MVTec-AD (mean and std of inference time over all 1725 test samples, and mean and std of time to populate the memory bank for each object) for different shots (\cref{subtab:shots}), preprocessing choices (\cref{subtab:preprocess}), sample resolutions (\cref{subtab:resolution}) and model sizes (\cref{subtab:model_size}). All times are reported in seconds, measured on a single NVIDIA A40 with GPU warmup and CUDA kernel synchronization. The default setting is 1-shot, agnostic preprocessing, model size S, and resolution 448 (\grayuline{underlined} for reference).}
    \label{tab:runtime}
    \vspace{-1mm}
    \def\tabsizeruntime{0.35}
        \begin{minipage}{\tabsizeruntime\textwidth}
        \centering
            \subcaptionbox{Runtime in dependence of shots. \label{subtab:shots}}{
            \begin{tabular}[t]{@{}ccc@{}}
            \toprule
            Shots               & \multicolumn{1}{c}{Inference} & Memory Bank \\ \midrule
            \grayuline{1}                   & 0.060\std{0.012}                   & \hphantom{0}0.52\std{0.04\hphantom{0}}   \\
            2                   & 0.059\std{0.012}                   & \hphantom{0}0.85\std{0.07\hphantom{0}}   \\
            4                   & 0.059\std{0.012}                   & \hphantom{0}1.58\std{0.08\hphantom{0}}   \\
            8                   & 0.060\std{0.011}                   & \hphantom{0}3.05\std{0.17\hphantom{0}}   \\
            16                  & 0.063\std{0.012}                   & \hphantom{0}6.02\std{0.39\hphantom{0}}   \\ \arrayrulecolor{lightgray} \midrule \arrayrulecolor{black}
            full \tiny{(masking only)}  & 0.067\std{0.010}           & 16.62\std{4.43\hphantom{0}}  \\
            full \tiny{(agnostic)} & 0.130\std{0.044}                & 93.72\std{20.56} \\ \bottomrule
            \end{tabular}
            }
        \end{minipage} \hspace{1cm}
        \begin{minipage}{\tabsizeruntime\textwidth}
        \vspace{-4.5mm}
        \centering
            \subcaptionbox{Runtime in dependence of preprocessing choices. \label{subtab:preprocess}}{
            \begin{tabular}[t]{@{}cccc@{}}
            \toprule
            Mask? & Rotate? & \multicolumn{1}{c}{Inference} & Memory Bank \\ \midrule
            no   & no       & 0.055\std{0.011}                   & 0.17\std{0.03}   \\
            no   & yes      & 0.055\std{0.012}                   & 0.51\std{0.05}   \\
            yes  & no       & 0.068\std{0.012}                   & 0.18\std{0.03}   \\ 
            yes  & yes      & 0.067\std{0.012}                   & 0.51\std{0.04}   \\ 
            \arrayrulecolor{lightgray} \midrule \arrayrulecolor{black}
            \multicolumn{2}{c}{informed} & 0.059\std{0.016}                   & 0.22\std{0.11}   \\  
            \multicolumn{2}{c}{\grayuline{agnostic}} & 0.060\std{0.012}                   & 0.52\std{0.04}   \\ \bottomrule
            \end{tabular}
            }
        \end{minipage}
        
        \vspace{2mm}
        
        \begin{minipage}{\tabsizeruntime\textwidth}
            \centering
            \subcaptionbox{Runtime in dependence of image resolution. \label{subtab:resolution}}{
            \begin{tabular}[t]{@{}ccc@{}}
            \toprule
            Resolution & \multicolumn{1}{c}{Inference} & Memory Bank \\ \midrule
            224        & 0.043\std{0.010}                   & 0.31\std{0.04}   \\
            \grayuline{448}        & 0.060\std{0.012}                   & 0.52\std{0.04}   \\
            672        & 0.086\std{0.014}                   & 0.75\std{0.06}   \\
            896        & 0.141\std{0.021}                   & 1.26\std{0.05}   \\ \bottomrule
            \end{tabular} 
            }
        \end{minipage} \hspace{1cm} 
        \begin{minipage}{\tabsizeruntime\textwidth}
            \centering
            \subcaptionbox{Runtime in dependence of model size. \label{subtab:model_size}}{
            \begin{tabular}[t]{@{}ccc@{}}
            \toprule
            Model Size & \multicolumn{1}{c}{Inference} & Memory Bank \\ \midrule
            \grayuline{S \tiny{(21 M)}}   & 0.060\std{0.012}                   & 0.52\std{0.04}   \\
            B \tiny(86 M)   & 0.084\std{0.021}                   & 0.77\std{0.05}   \\
            L \tiny(300 M)  & 0.141\std{0.029}                   & 1.24\std{0.06}   \\
            G \tiny(1,100 M) & 0.306\std{0.034}                   & 2.67\std{0.14}   \\ \bottomrule
            \end{tabular}
            }
        \end{minipage}
\end{table*}

\paragraph{Runtime analysis} 
The design choices in our pipeline influence the run time of the proposed method.
To see the potential effects we measure the inference time per sample on MVTec-AD in various scenarios together with the time to set up the memory bank $\mathcal{M}$. The runtime was assessed utilizing a single NVIDIA A40 GPU, consistently employed throughout all experiments detailed in this paper (each experiment can be executed on a single NVIDIA A40 GPU). The runtime is measured with GPU warmup and CUDA kernel synchronization for a fair comparison.

The results are given in \Cref{tab:runtime}. 
The average inference time per sample with \method{}-S (448) amounts to approximately 60ms in the few-shot regime ($\approx$ 16.7 samples/s), and only moderately increases with a larger memory bank (to 67ms ($+11\%$) for the full-shot scenario without augmentations).
Compared to SOTA competitors in the one-shot regime, this is at least one order of magnitude faster (see \Cref{fig:detection_vs_inferencetime}).
Without any preprocessing steps, the 1-shot inference time amounts to approximately 55ms per sample ($\approx$ 18 samples/s).
When applying both masking and rotations, the runtime increases from 55ms to 67ms, a moderate increase of roughly 23\% (compared to the scenario without any preprocessing steps). The informed scenario can reduce the time to build the memory bank, but inference time is only affected for larger sample sizes.
As expected, higher resolutions and larger architectures lead to increased runtimes.

\subsection{Scoring} 
\label{App:Ablation-Scoring}

\begin{table*}[!hbt]
\centering
\caption{\textbf{Effect of aggregation statistics $q$} on the detection performance on MVTec-AD and VisA, evaluated for \method{}-S (448). $\mathcal{D}$ denotes the set of patch distances to the nominal memory bank, $H_{0.01}(\mathcal{D})$ the 1\% highest entries thereof, and $\mathcal{A}$ denotes the anomaly map derived from $\mathcal{D}$ (anomaly scores for each pixel, after upsampling and smoothing, see \Cref{sec:Method}). All results in \%.}
\resizebox{0.93\textwidth}{!}{
\begin{tabular}{@{}ccccccccccccc@{}}
\toprule
\multicolumn{2}{r}{Scoring}       & \multicolumn{3}{c}{$q = \mathrm{mean}(H_{0.01}(\mathcal{D}))$} &  & \multicolumn{3}{c}{$q= \mathrm{max}(\mathcal{\mathcal{D}})$} &  & \multicolumn{3}{c}{$q = \mathrm{max}(\mathcal{A})$} \\ \cmidrule(l){3-13} 
\multicolumn{2}{r}{Shots} & AUROC              & F1-max             & AP                &  & AUROC              & F1-max             & AP                 &  & AUROC           & F1-max          & AP              \\ \midrule
\parbox[t]{2mm}{\multirow{5}{*}{\rotatebox[origin=c]{90}{MVTec-AD}}}
&1 & 96.5\std{0.4} & 96.0\std{0.2} & 98.1\std{0.3}  &  & 95.0\std{0.5} & 94.7\std{0.2} & 97.4\std{0.4}      &  & 94.9\std{0.7} & 94.6\std{0.6} & 97.5\std{0.4}   \\
&2 & 96.7\std{0.8} & 96.5\std{0.4} & 98.1\std{0.7}  &  & 95.5\std{1.0} & 95.3\std{0.5} & 97.4\std{0.6}      &  & 95.0\std{1.3} & 94.8\std{0.8} & 97.4\std{0.9}   \\
&4 & 97.6\std{0.1} & 97.0\std{0.3} & 98.4\std{0.3}  &  & 96.6\std{0.2} & 96.0\std{0.1} & 97.8\std{0.4}      &  & 96.3\std{0.4} & 95.7\std{0.2} & 98.1\std{0.3}   \\
&8 & 98.0\std{0.1} & 97.4\std{0.1} & 99.0\std{0.2}  &  & 97.2\std{0.1} & 96.5\std{0.4} & 98.6\std{0.1}      &  & 97.0\std{0.0} & 96.3\std{0.1} & 98.6\std{0.1}   \\
&16 & 98.3\std{0.1} & 97.7\std{0.2} & 99.3\std{0.0} &  & 97.6\std{0.2} & 96.9\std{0.3} & 98.9\std{0.1}      &  & 97.4\std{0.1} & 96.6\std{0.2} & 98.8\std{0.1}   \\ \midrule
\parbox[t]{2mm}{\multirow{5}{*}{\rotatebox[origin=c]{90}{VisA}}}
&1 & 85.6\std{1.5} & 83.1\std{1.1} & 86.6\std{1.3}  &  & 82.4\std{1.9} & 81.4\std{1.0} & 84.0\std{1.7}      &  & 80.5\std{1.3} & 80.6\std{0.8} & 82.1\std{0.9}   \\
&2 & 88.3\std{1.8} & 84.8\std{1.2} & 89.2\std{1.3}  &  & 85.1\std{1.8} & 82.5\std{1.3} & 86.4\std{1.1}      &  & 83.3\std{1.8} & 82.2\std{1.1} & 84.5\std{1.5}   \\
&4 & 91.3\std{0.8} & 87.5\std{1.0} & 91.8\std{0.7}  &  & 88.4\std{0.3} & 84.9\std{0.6} & 89.0\std{0.4}      &  & 86.8\std{1.4} & 84.3\std{0.9} & 87.6\std{1.3}   \\
&8 & 92.6\std{0.1} & 88.6\std{0.2} & 92.9\std{0.2}  &  & 90.0\std{0.3} & 86.3\std{0.2} & 90.6\std{0.5}      &  & 88.8\std{0.5} & 85.3\std{0.3} & 89.7\std{0.3}   \\
&16 & 93.8\std{0.1} & 89.9\std{0.3} & 94.2\std{0.3} &  & 91.6\std{0.5} & 87.2\std{0.7} & 92.2\std{0.4}      &  & 90.5\std{0.3} & 86.8\std{0.5} & 91.5\std{0.4}   \\ \bottomrule
\end{tabular}
} 
\label{tab:ablation_q}
\end{table*}

In \Cref{sec:Method} we investigate different ways of aggregating the anomaly scores $\mathcal{D}$ on patch-level (in our case, distances to the nominal memory bank $\mathcal{M}$) to an image-level anomaly score via a statistic $q$.
Note that the segmentation results are therefore not affected by the choice of $q$. 
A standard choice is upsampling the patch distances of lower resolution to the full resolution of the test image using bilinear interpolation, then applying Gaussian smoothing operation (we follow \cite{roth2022towards} and set $\sigma=4.0$), to obtain an anomaly map $\mathcal{A}$, and set $q = \max{(\mathcal{A})}$.
We also evaluate two potential alternatives of $q$, our default choice $q = \mathrm{mean}(H_{0.01}(\mathcal{D}))$ (mean of the 1\% highest entries in $\mathcal{D}$) and $q=\max(\mathcal{D})$.

The results in \Cref{tab:ablation_q} show that just the maximum patch distance leads to already good results while upsampling and smoothing the patch distances giving slightly weaker results (maximum of $\mathcal{A}$).

Taking the mean of all patch distances above the 99\% quantile ($q = \mathrm{mean}(H_{0.01}(\mathcal{D}))$) improves above the standard choice.
We did not optimize over the percentile and instead fixed it to 99\%.
Typically, the number of patches per image ranges between 200 and 1000 (depending on resolution and masking), such that 2 between 10 patches are considered.
For specific products/objects and expected anomaly types, other statistics $q$ might be (more) suitable.


    

\subsection{Architecture Size} 
\label{App:Ablation-Size}

\begin{figure*}[!bth]
    \centering
    
    \begin{subfigure}[t]{0.49\linewidth}
    \centering
    \includegraphics[width=\linewidth]{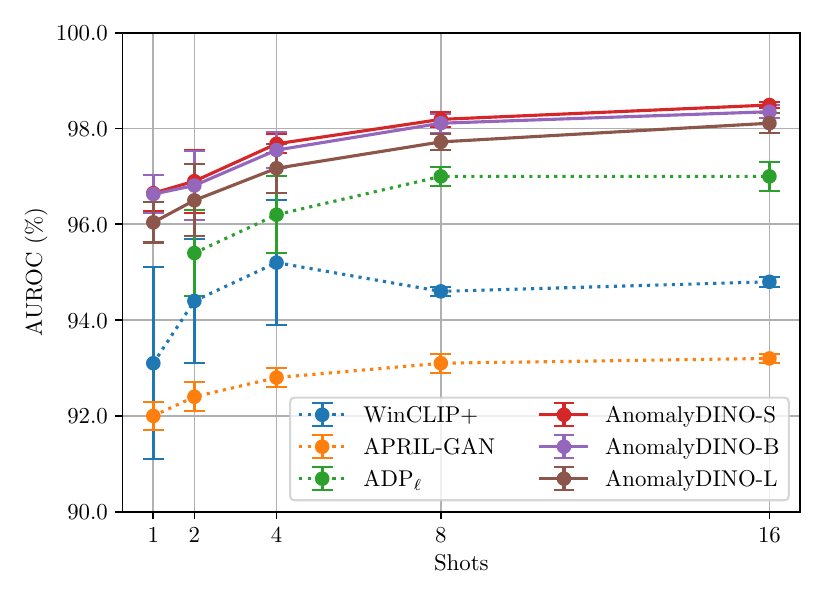} 
    \vspace{-5mm}
    \caption{Detection AUROC for MVTec-AD.}
    \end{subfigure}
    \begin{subfigure}[t]{0.49\linewidth}
    \centering
    \includegraphics[width=\linewidth]{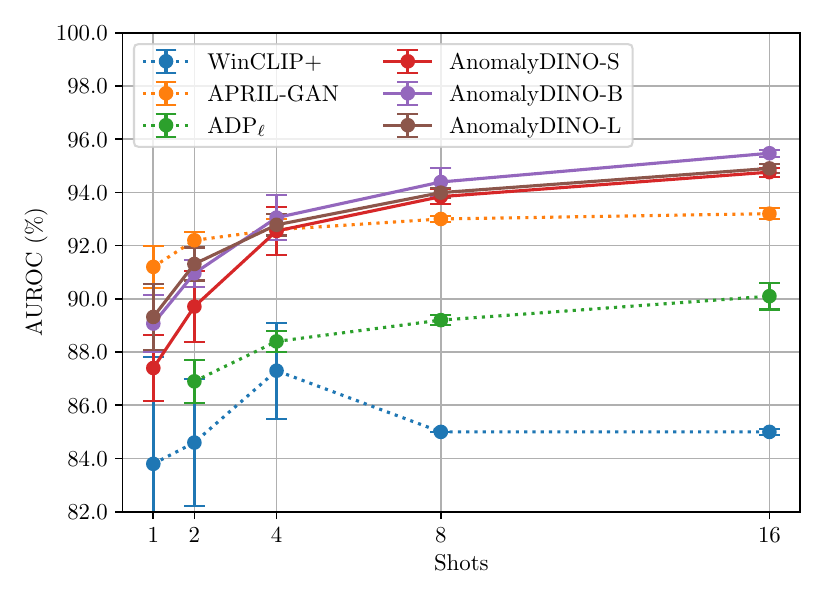}
    \vspace{-5mm}
    \caption{Detection AUROC for VisA.}
    \end{subfigure}
    
    \vspace{-1mm}
    \caption{\textbf{Effect of model size} on detection AUROC for MVTec-AD and VisA (mean and std over three seeds).
    The image resolution of \method{} is set to 672.
    Note, that the results for WinCLIP+ for 8 and 16 shots are those of the WinCLIP re-implementation \cite{kwak2024few}.}
    \label{fig:ablation_model_size}
\end{figure*}

DINOv2 is available in different distillation sizes (S, B, L, and G).
This section analyzes the implications of choosing different backbone sizes for \method{}.
The comparison including the best competing methods is depicted in \Cref{fig:ablation_model_size}.
We see that different architecture sizes have indeed an effect on the image-level AUROC.

On MVTec-AD, we see that \method{}-S performs best, followed by the next larger model \method{}-B (which might contrast the common belief that larger models always perform better).
In particular, all architecture sizes outperform the closest competitor (ADP$_\ell$).

Regarding VisA, larger architectures slightly outperform our default setting, \method{}-S.
All architecture sizes are on par with APRIL-GAN at $k=4$, but outperform all competitors for $k > 4$.
\Cref{fig:ablation_model_size} also showcases that the performance of the proposed method scales preferably with the number of reference samples.

\begin{figure*}[!bht]
    \centering
    
    \begin{subfigure}[t]{0.49\linewidth}
    \centering
    \includegraphics[width=1.0\linewidth]{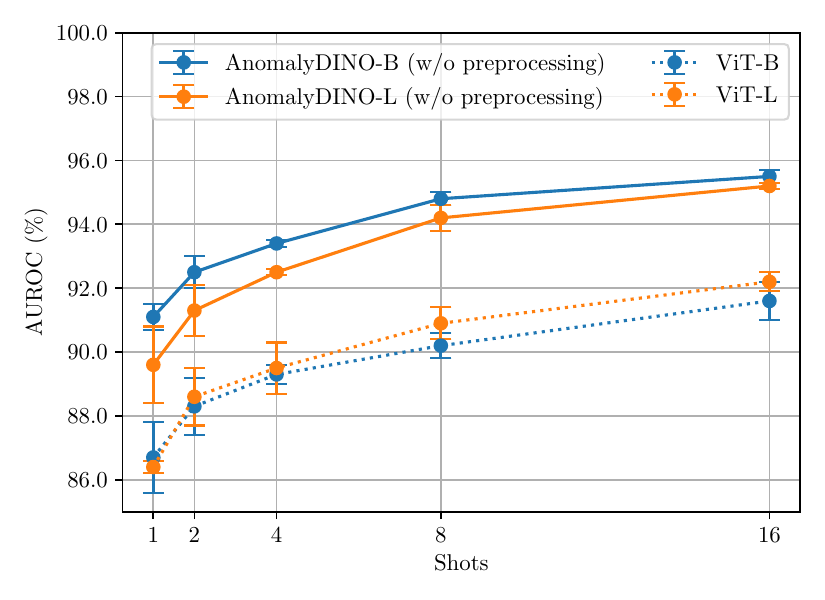} 
    \vspace{-5mm}
    \caption{Detection AUROC for MVTec-AD.}
    \end{subfigure}
    \begin{subfigure}[t]{0.49\linewidth}
    \centering
    \includegraphics[width=\linewidth]{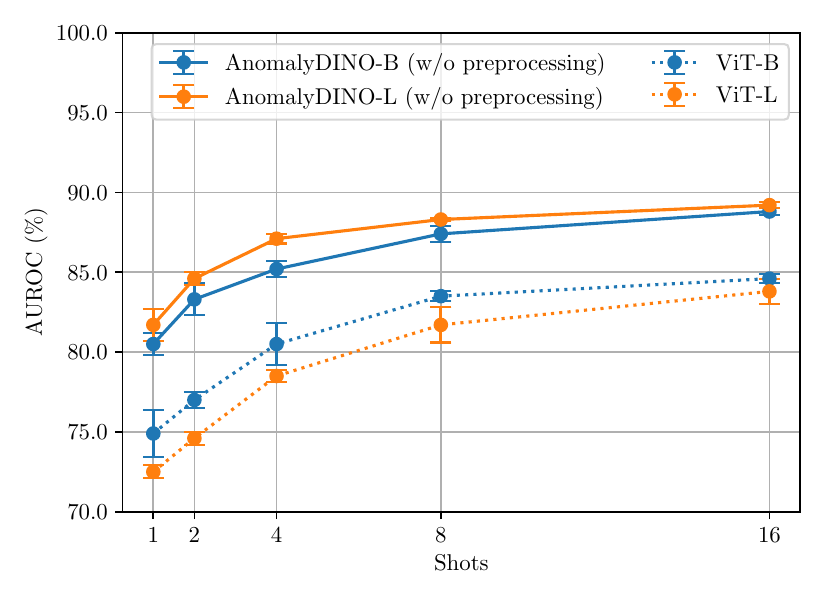}
    \vspace{-5mm}
    \caption{Detection AUROC for VisA.}
    \end{subfigure}
    
    \vspace{-1mm}
    \caption{\textbf{Effect of backbone choice} DINOv2 (trained in a self-supervised fashion) compared to ViT (supervised training on ImageNet)
    on detection AUROC for MVTec-AD and VisA (mean and std over three seeds).
    For better comparability, image resolution of \method{} is 224 for all models.}
    \label{fig:ablation_backbone}
\end{figure*}

\subsection{Backbone Choice}
\label{app:ablation-backbone}

In our experiments, we find that DINOv2 provides excellently suited features for few-shot AD.
This is already evident in \cref{fig:detection_vs_inferencetime}, here we investigate the effect on the detection performance on MVTec-AD and VisA in more detail.

We compare the performance in few-shot AD of ViT-B and ViT-L pre-trained on ImageNet to that of AnomalyDINO-B and AnomalyDINO-L (utilizing DINOv2 in distillation sizes B and L). 
To ensure comparability, we deactivate any pre- and postprocessing for \method{} (no augmentations, no zero-shot masking) and set the image resolution to 224 for all models. 

The results are depicted in \Cref{fig:ablation_backbone}.
We observe that DINOv2 significantly improves the detection performance compared to those of the ViT pre-trained on the image classification tasks, giving a performance gain of at least $+4\%$ AUROC on MVTec-AD and VisA. This demonstrates that the features extracted by DINOv2 are better suited compared to those of ImageNet-pretrained ViT-\{B/L\}. 

In \Cref{sec:Method,sec:Experiments} we demonstrate that further (substantial) improvements are possible with suitable pre- and postprocessing like augmentations and zero-shot masking (which is not possible based on the features from supervised features) and that the image resolution (or the effective patch size) can also greatly boost performance. See also \cref{fig:ablation_model_size} for the performance in dependence of the model size.

\section{Extending \method{} to the Batched Zero-Shot Setting}
\label{App:Batched-0-Shot}

We extend the proposed method to the \textit{batched} zero-shot setting. Recall, that in this setting all test samples $X_\mathrm{test}$ are provided (or at least a sufficiently large batch), but \textit{no} (labeled) training or reference samples.
The underlying (and necessary) assumption to meaningfully predict anomalies solely based on test samples, is that the majority of samples (or in our case, patches) at test time are from the nominal data distribution (see e.g., \cite{li2024zero}).

We need to alter our method, outlined in \Cref{sec:Method}, only slightly to adapt it to the batched setting.
We score a test sample $\mathbf{x}^{(j)} \in X_\mathrm{test}$ in comparison to all remaining test sample $X_{\operatorname{test}} \setminus \{\mathbf{x}^{(j)}\}$, following the idea of mutual scoring \cite{Li2024MuSc}. 
For each test sample $\mathbf{x}^{(j)} \in X_\mathrm{test}$ we therefore collect all patch representations not belonging to $\mathbf{x}^{(j)}$ in a memory bank, again utilizing DINOv2 as patch-level feature extractor $f$,

\begin{equation}
    \label{eq:memory_bank_batched}
    \mathcal{M}_{j} := \hspace{-7mm} \bigcup_{\mathbf{x}^{(i)} \in X_{\operatorname{test}} \setminus \{\mathbf{x}^{(j)}\}} \hspace{-7mm}  \bigl\{ \mathbf{p}_m \mid f(\mathbf{x}^{(i)}) = (\mathbf{p}_1, \dots, \mathbf{p}_n), \, m \in [n] \bigr\}  \enspace . 
\end{equation}

We need to infer anomaly scores for each patch representation $\mathbf{p}_\mathrm{test}$ of $\mathbf{x}^{(j)}$. 
We could again assess the distances between $\mathbf{p}_\mathrm{test}$ and $\mathcal{M}_{j}$ based on the distance to the nearest neighbor, as done in \Cref{eq:NN-distance}.
Note, however, that $\mathcal{M}_{j}$ may now contain nominal \textit{and} anomalous patches. Thus, the nearest neighbor approach is not suitable anymore: the nearest neighbor in $\mathcal{M}_{j}$ for (the representation of) an anomalous patch might also be abnormal, and the resulting distance therefore not informative.

\begin{figure}[!thb]
    \centering
    \includegraphics[width=1.0\linewidth]{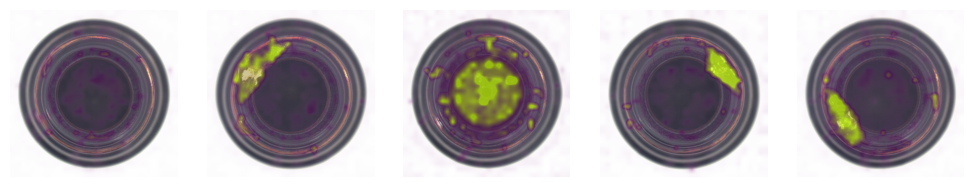} 
    
    \includegraphics[width=1.0\linewidth]{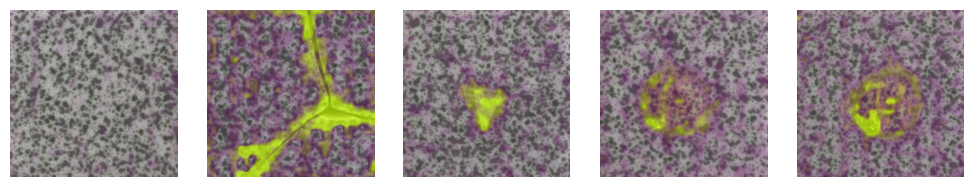} 
    
    \includegraphics[width=1.0\linewidth]{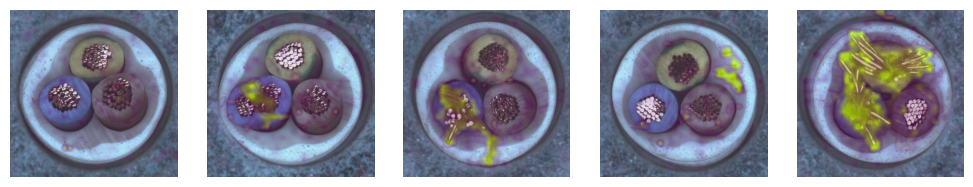} 
    
    \includegraphics[width=1.0\linewidth]{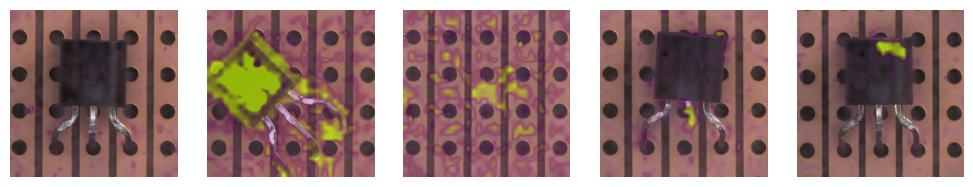} 
    
    \includegraphics[width=1.0\linewidth]{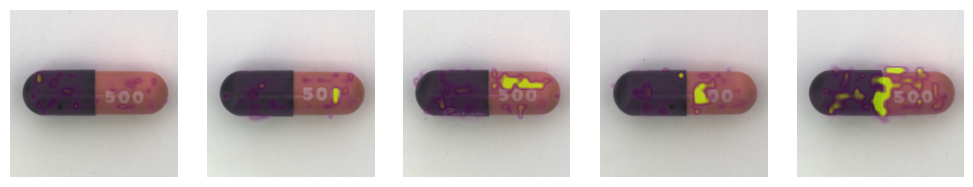} 
    
    \includegraphics[width=1.0\linewidth]{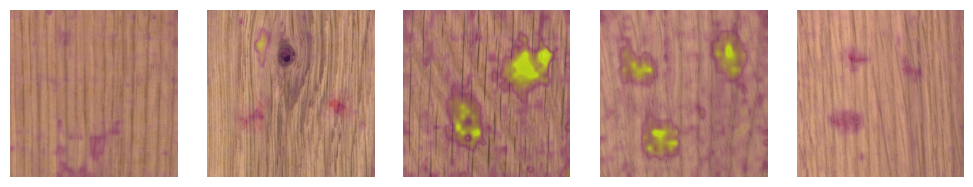} 
    \caption{\textbf{Anomaly maps for the batched zero-shot setting} on MVTec-AD.
    The left-most sample in each category is a `good' test sample for reference, followed by four randomly picked samples with anomalies.}
    \label{fig:examples_batched-0-shot}
\end{figure}

A simple solution, based on the assumption that the majority of patches are nominal, is to replace the nearest neighbor with a suitable aggregation statistic over the distribution of patches distances in $\mathcal{M}_{j}$
\begin{equation}
   \mathcal{D}(\mathbf{p}_\mathrm{test}, \mathcal{M}_j) := \big\{d(\mathbf{p}_\mathrm{test}, \mathbf{p}) \mid \mathbf{p} \in \mathcal{M}_j \big\} 
   \enspace,
\end{equation}
where we again use the cosine distance $d$, defined in \Cref{eq:cosine_distance}.
Specifically, we can again make use of the tail value at risk---now for the lowest quantile as we are interested in the tail behavior of the distribution of distances to the nearest neighbors---to derive a patch-level anomaly score
\begin{equation}
    s(\mathbf{p}_\mathrm{test}) := 
        \mathrm{mean}\big( L_\alpha
            \big(\mathcal{D}(\mathbf{p}_\mathrm{test}, \mathcal{M}_j)\big)\big) \enspace,
\end{equation}
where $L_{\alpha}(\mathcal{D})$ contains the values below the $\alpha$\ quantile of $\mathcal{D}$. We set $\alpha = 0.1\%$ as anomalous patches are rare by assumption, and $\mathcal{D}(\mathbf{p}_\mathrm{test}, \mathcal{M}_j)$ large enough to accurately estimate the tail of the distribution.\footnote{For MVTec-AD the total number of test patches extracted by DINOv2 (ViT-S) at a resolution of 448 range between 43.008 and 171.008 per category, and for VisA between 163.200 and 334.464.}
The image-level score  $s(\mathbf{x}_\mathrm{test})$ is again given by aggregating the patch-level anomaly score following \Cref{eq:meantop1p}.
Computation of the cosine distances and the proposed aggregation statistics can be effectively implemented as matrix operations on GPU such that the batched zero-shot inference time for AnomalyDINO amounts to roughly 60 ms/sample for MVTec-AD at a resolution of 448 (again measured on an NVIDIA A40 GPU).
Some resulting anomaly maps in the batched zero-shot setting are depicted in \Cref{fig:examples_batched-0-shot}.

\section{Broader Impacts}

Advancing few-shot visual anomaly detection methods can offer various benefits by enhancing manufacturing quality control through the rapid identification of defects with minimal nominal examples, which improves efficiency, reduces waste, and improves overall safety in the product lifecycle. 
Similar positive benefits can be expected outside the industrial domain, e.g., for healthcare diagnostics or environmental monitoring.
It is, however, essential to recognize the shortcomings of automated anomaly detection systems. We believe that simpler methods can be adapted more quickly, monitored more effectively, and are therefore more reliable. 
In this context, awareness of the risks of overreliance must be heightened (see \Cref{App:Limitations} for identified failure cases of the proposed method).
In addition, strong visual anomaly detectors could also lead to potentially malicious or unintended uses.
To address these concerns, including potential privacy infringements and possible socioeconomic impacts of automation, strategies such as establishing robust data governance, 
and implementing strict privacy protections are essential. 
Additionally, investing in workforce development can help manage the socioeconomic effects of automation and leverage the full potential of strong visual anomaly detection.

\end{document}